\documentclass[10pt,journal,compsoc]{IEEEtran}
\ifCLASSOPTIONcompsoc
  \usepackage[nocompress]{cite}
\else
  \usepackage{cite}
\fi
\ifCLASSINFOpdf
  \usepackage[pdftex]{graphicx}
\else
\fi
\usepackage{amsmath}
\usepackage{amssymb}

\ifCLASSOPTIONcompsoc
  \usepackage[caption=false,font=footnotesize,labelfont=sf,textfont=sf]{subfig}
\else
  \usepackage[caption=false,font=footnotesize]{subfig}
\fi
\usepackage{multirow}
\usepackage{pifont}% http://ctan.org/pkg/pifont
\newcommand{\cmark}{\ding{51}}%
\newcommand{\xmark}{\ding{55}}%

\usepackage[caption=false]{subfig}
\usepackage{caption}
\usepackage{dblfloatfix}
\usepackage{enumitem}
\usepackage{xcolor}

\begin{document}
\title{Free-HeadGAN: Neural Talking Head Synthesis with Explicit Gaze Control}

\author{Michail~Christos~Doukas,
        Evangelos~Ververas,
        Viktoriia~Sharmanska,
        and~Stefanos~Zafeiriou,~\IEEEmembership{Member,~IEEE}% <-this % stops a space
\IEEEcompsocitemizethanks{\IEEEcompsocthanksitem The authors are with the Department of Computing, Imperial College London, London SW7 2AZ, U.K. E-mail:
\{michail-christos.doukas16, e.ververas16, sharmanska.v, s.zafeiriou\}@imperial.ac.uk.
\IEEEcompsocthanksitem M. C. Doukas, E. Ververas and S. Zafeiriou are also with Huawei Technologies London, U.K.
\IEEEcompsocthanksitem V. Sharmanska is also with the Department of Computing, University of Sussex, London SW7 2AZ, U.K. }% <-this % stops an unwanted space
% TODO: Uncomment
%\thanks{Manuscript received month dd, yyyy; revised month dd, yyyy.}
\thanks{}
}


% TODO: Uncomment
%\markboth{Journal of \LaTeX\ Class Files,~Vol.~XX, No.~X, month~yyyy}
\markboth{}
{Doukas \MakeLowercase{\textit{et al.}}: Free-HeadGAN: Model-free Neural Talking Head Synthesis}



\IEEEtitleabstractindextext{%
\begin{abstract}
We present Free-HeadGAN, a person-generic neural talking head synthesis system. We show that modeling faces with sparse 3D facial landmarks are sufficient for achieving state-of-the-art generative performance, without relying on strong statistical priors of the face, such as 3D Morphable Models. Apart from 3D pose and facial expressions, our method is capable of fully transferring the eye gaze, from a driving actor to a source identity. Our complete pipeline consists of three components: a canonical 3D key-point estimator that regresses 3D pose and expression-related deformations, a gaze estimation network and a generator that is built upon the architecture of HeadGAN. We further experiment with an extension of our generator to accommodate few-shot learning using an attention mechanism, in case more than one source images are available. Compared to the latest models for reenactment and motion transfer, our system achieves higher photo-realism combined with superior identity preservation, while offering explicit gaze control. 
\end{abstract}

}

\maketitle

\IEEEdisplaynontitleabstractindextext

% \fi
\IEEEpeerreviewmaketitle

\IEEEraisesectionheading{\section{Introduction}\label{sec:introduction}}

\IEEEPARstart{G}{enerating} photo-realistic images or videos of human faces has lately become a very popular computer vision topic, with a remarkable amount of research and discussion revolving around it. Aside from considerable advancements accomplished by the computer graphics community \cite{thies2015realtime, face2face, thies2018headon}, deep neural networks and more specifically \textit{Generative Adversarial Neural Networks (GANs)} \cite{goodfellow2014generative} have made a significant contribution towards this direction. For instance, \textit{StyleGAN2} \cite{stylegan2} has shown incredible results on unconditional faces synthesis.%, truly indistinguishable from real data.

Neural talking head synthesis, here also referred to as \textit{reenactment}, is the task where a reference (source) image or video is manipulated or re-animated to match with the facial expressions and head poses performed by a driving (target) actor. Reenactment systems have numerous applications in social media, teleconference, image and video editing, as well as virtual reality and games. They can be further used for facial video compression and reconstruction \cite{wang2021facevid2vid}.

\begin{figure}[h!]
\centering
\includegraphics[width=0.44\textwidth]{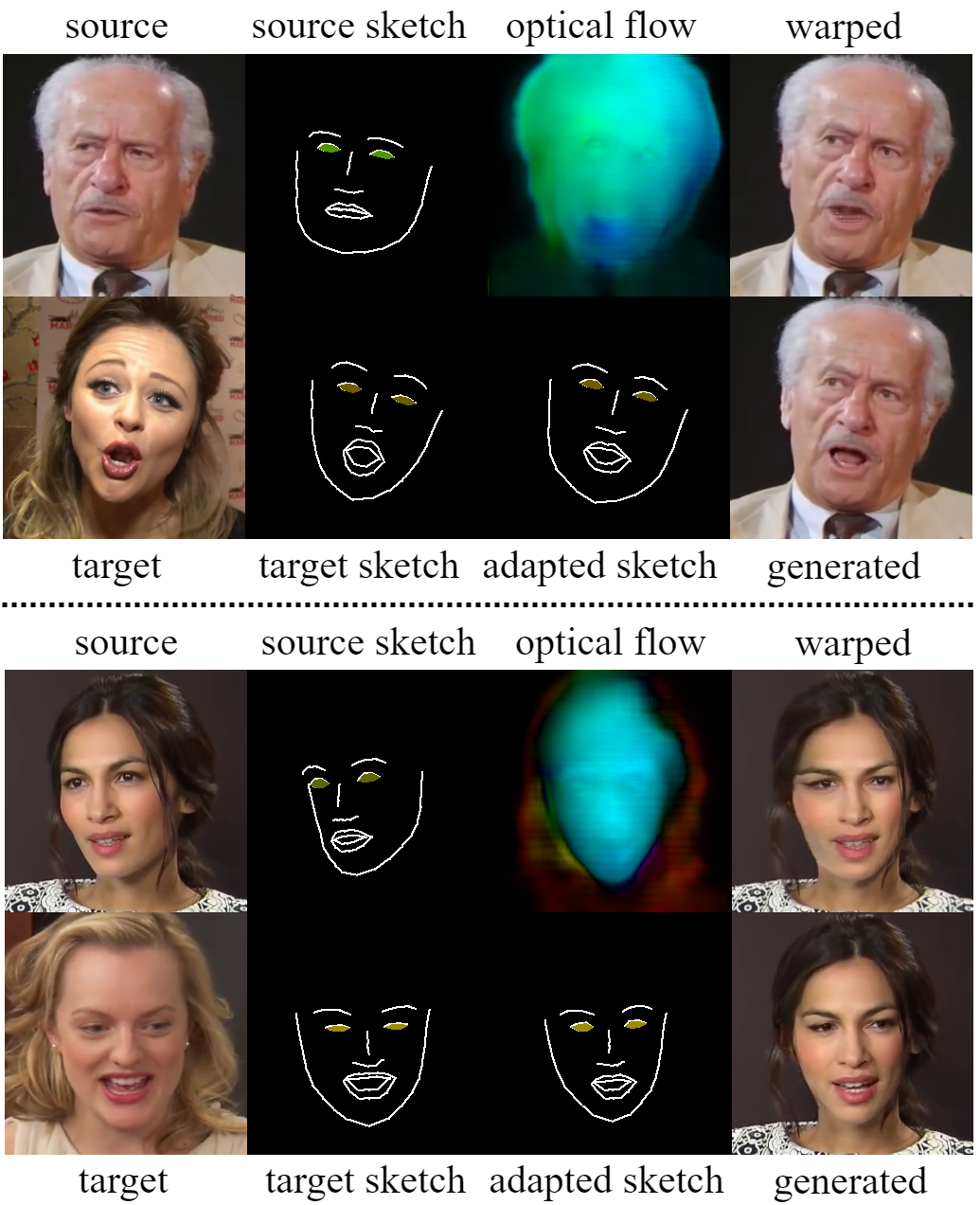}
\caption{Given a source and a target image, first we compute a flow field to warp the source image, according to the target pose, expression and gaze. Then, a rendering network generates the final image based on the warped image (and warped visual features). We condition synthesis on sketches of 3D key-points. We preserve identities, by adapting the target key-points to the facial shape of the source, using a canonical 3D key-point estimator. We control gaze in the generated faces by color coding the interior of the eye cavities in sketches with the gaze direction angles.
%In order to control gaze in the generated faces, we color code the interior of the eye cavities in sketches with the eye rotation angles of the target, after estimating their gaze.
}
\label{fig:canonical}
\end{figure}

Learning-based reenactment methods are classified either as \textit{person-specific} or \textit{person-generic}. On the one hand, person-specific approaches \cite{kim2018deep,head2head2020,head2headpp} are based on identity-specific neural networks. Despite the impressive visual results, these networks require re-training using a long video of the particular source actor we wish to reenact. On the other hand, person-generic systems \cite{Zakharov2019FewShotAL, fsvid2vid, bilayer, MarioNETte:AAAI2020, X2Face, Siarohin_2019_CVPR, FOMM, wang2021facevid2vid, doukas2020headgan} have the ability to adapt to the source identity during inference, given only a few reference images of the subject, even a single one. Although very promising, in general such models generate samples of lesser quality, when compared with their identity-specific counterparts.

Another important aspect of talking head synthesis relates to face modeling. A common face representation is 2D sparse landmarks (68 points), as used in \cite{Zakharov2019FewShotAL, fsvid2vid, bilayer}. More recent methods, such as \cite{MarioNETte:AAAI2020} and \cite{Geng2018WarpguidedGF} use 3D landmarks instead. Among other benefits, the selection of a three-dimensional representation enables \textit{free-view control}, as we can manually intervene in the head pose to be generated. Other widely used facial representations are dense 3D facial meshes combined with statistical priors of the face, such as \textit{3D Morphable Models (3DMMs)} \cite{3dmms, 3dmmsbooth, Booth, Booth_2016_CVPR}, which are adopted for reenacting heads in \cite{kim2018deep, head2head2020, doukas2020headgan}. This modeling enables to disentangle shape (identity) from expression-related deformations, and therefore helps to tackle the \textit{identity preservation problem} in cross-identity motion transfer, which is neither straightforward nor trivial with sparse landmarks. In a quite different line of work, Siarohin \textit{el al.} \cite{Siarohin_2019_CVPR, FOMM} attempt to solve the most general problem of motion transfer for arbitrary objects, proposing a \textit{model-free} method based on unsupervised 2D key-point detection. Following up, Wang \textit{el al.} \cite{wang2021facevid2vid} present a model that focuses on faces and learns to detect \textit{canonical} 3D key-points without supervision, as a way to disengage identity from expression and pose.

Given that they deal with motion transfer, most recent systems are flow-based \cite{MarioNETte:AAAI2020, Geng2018WarpguidedGF, X2Face, Siarohin_2019_CVPR, FOMM, wang2021facevid2vid, doukas2020headgan}, meaning that they make use of optical flow to warp the source image(s) or visual features into the target facial pose and expression, as a first step prior to image synthesis. The aforementioned model-free approaches \cite{Siarohin_2019_CVPR, FOMM, wang2021facevid2vid} are able to learn important key-points for motion, as they approximate optical flow in areas near these points. For example, \cite{FOMM} and \cite{wang2021facevid2vid} use a \textit{first order approximation of motion}. On the contrary, methods such as \cite{X2Face, MarioNETte:AAAI2020, doukas2020headgan} learn a \textit{dense optical flow} in all spatial locations, without resorting to approximations. 

In this work, we do not use a first order approximation of motion, but rather use a neural network that learns to predict dense optical flow, similarly to \textit{HeadGAN} \cite{doukas2020headgan}. We propose \textit{Free-HeadGAN}, a system based on the architecture of \cite{doukas2020headgan} which is "free" from statistical priors of faces like 3DMMs. In order to overcome the identity preservation obstacles in reenactment and inspired from \cite{wang2021facevid2vid}, we propose a network for regressing 3D key-points along with head pose and expression deformations. Different from \cite{wang2021facevid2vid} our method does not predict canonical key-points directly. Instead, we compute them based on the estimated 3D points after removing pose and expression. Moreover, we supervise our key-point detector with pseudo ground-truth 3D facial landmarks extracted with \textit{RetinaFace} \cite{RetinaFace}. In this way, we ensure that key-points are always placed on meaningful parts of the face, avoiding cases where no key-points are assigned to important regions such as the eyes and mouth, as it commonly happens with unsupervised models \cite{Siarohin_2019_CVPR, FOMM, wang2021facevid2vid}. Lastly, we propose a gaze estimation network that makes explicit eye gaze transfer possible. The contributions of our work can be summarised in the following:
\begin{itemize}[leftmargin=*]
    \item We release \textit{HeadGAN} \cite{doukas2020headgan} from its 3DMM priors, using sparse 3D landmarks. We achieve comparable and in many cases improved results, as suggested by our qualitative and quantitative experiments.
    \item We design a module that performs disentanglement of identity, expression and pose through the computation of canonical 3D key-points.
    \item We propose a network that regresses 3D meshes of the eyes. We use these meshes to obtain the direction of gaze, which is then used to condition image synthesis. To the best of our knowledge, we deliver the first person-generic reenactment system with explicit gaze control.
    \item We show a N-shot extension of our framework.
\end{itemize}

\section{Background}\label{sec:Background}

\begin{table*}[!t]
\renewcommand{\arraystretch}{1.15}
\caption{Checklist of key features and design choices of state-of-the-art head synthesis systems that support full head animation and  reenactment}
\label{table:background}
\centering
\begin{tabular}{|c||c|c|c|c|c|c|c|c|}
\hline
\multirow{2}{*}{Method} & Person & \multirow{2}{*}{N-shot} & Face & Free-view & Identity & Gaze & Warping & \multirow{2}{*}{Sequential} \\
 & generic & & modeling & control & preservation & control & flow & \\
\hline
\textit{DVP} \cite{kim2018deep} & \xmark & - & 3DMM & \cmark & \cmark & \cmark & \xmark & \xmark \\
\textit{Head2Head} \cite{head2head2020} & \xmark & - & 3DMM & \cmark & \cmark & \cmark & \xmark & \cmark \\
Zakharov \textit{et al.} \cite{Zakharov2019FewShotAL} & \cmark & \cmark & 2D landmarks & \xmark & \xmark & \xmark & \xmark & \xmark \\
\textit{Few-shot vid2vid} \cite{fsvid2vid} & \cmark & \cmark & 2D landmarks & \xmark & \xmark & \xmark & \xmark & \cmark \\
Zakharov \textit{et al.} \cite{bilayer} & \cmark & \xmark & 2D landmarks & \xmark & \xmark & \xmark & \xmark & \xmark \\
\textit{MarioNETte} \cite{MarioNETte:AAAI2020} & \cmark & \cmark & 3D landmarks & \xmark & \cmark & \xmark & learned & \xmark \\
\textit{Warp-Guided GAN}s \cite{Geng2018WarpguidedGF} & \cmark & \xmark & 3D landmarks & \cmark & \xmark & \xmark & crafted & \xmark \\
\textit{X2Face} \cite{X2Face} & \cmark & \cmark & embeddings & \xmark & \xmark & implicit & learned & \xmark \\
\textit{Monkey-Net} \cite{Siarohin_2019_CVPR} & \cmark & \xmark & 2D key-points & \xmark & \xmark & implicit & approx. & \xmark \\
\textit{FOMM} \cite{FOMM} & \cmark & \xmark & 2D key-points & \xmark & \xmark & implicit & approx. & \xmark \\
\textit{face-vid2vid} \cite{wang2021facevid2vid} & \cmark & \xmark & 3D key-points & \cmark & \cmark & implicit & approx. & \xmark \\
\textit{HeadGAN} \cite{doukas2020headgan} & \cmark & \xmark & 3DMM & \cmark & \cmark & \xmark & learned & \xmark \\
\textit{Free-HeadGAN} & \cmark & \cmark & 3D landmarks & \cmark & \cmark & \cmark & learned & \xmark \\
\hline
\end{tabular}
\end{table*}

\subsection{GANs and Image Synthesis}

Since their introduction, \textit{GANs} \cite{goodfellow2014generative} have been extensively used both for unconditional and conditional \cite{mirza2014conditional} data synthesis. They have been successfully applied to various computer vision tasks, such as image-to-image \cite{pix2pix2017, CycleGAN2017, wang2018pix2pixHD, park2019SPADE}, video-to-video \cite{wang2018vid2vid, fsvid2vid} and audio-to-image translation \cite{Vougioukas, thies2020nvp}. In this work, we focus on the problem of conditional image synthesis and propose a GAN-based talking head synthesis system that achieves high photo-realism and improved identity preservation compared to current state-of-the-art approaches. 

\subsection{Talking Head Synthesis}

\subsubsection{Face Reenactment}

Most of the early attempts in human face synthesis focus on the task of \textit{face reenactment}, which aims to transfer the facial expressions performed in a driving video to the face of another person. The majority of face reenactment methods, i.e. \textit{Face2Face} \cite{face2face}, manipulate the source footage, by re-writing only the facial region of the source video stream, while keeping the remaining parts (e.g. hair, body, background) unchanged \cite{GVRTPT14, obama}. A first exception to those systems has been \textit{Bringing Portraits to Life} \cite{elor2017bringingPortraits}, which is able to slightly animate the entire head with the application of 2D warping on the source image.

\subsubsection{Full Head Reenactment}

\textit{Deep Video Portraits (DVP)}~\cite{kim2018deep} is allegedly one of the earliest learning-based reenactment methods, as it utilises an image translation network that manages to fully transfer the head motions of the driver to the source, including eye gaze. In a more recent work, \textit{Head2Head}~\cite{head2head2020, head2headpp} adopts a sequential video-based translation network that performs \textit{full head reenactment} while taking into consideration the temporal dynamics of talking faces. Both aforementioned methods rely on 3D information of faces and strong statistical priors, such as \textit{3DMMs} \cite{3dmms, 3dmmsbooth, Booth, Booth_2016_CVPR}. Despite their remarkable results, these \textit{person-specific} models are optimised on long videos of the source identity, being incapable of adapting and generalising to new identities without re-training.

On the other hand, \textit{person-generic} methods are able to adapt their generative process on the appearance of new source identities, even if they are not introduced to them during training. Numerous identity-agnostic talking head synthesis systems are 2D-based \cite{Zakharov2019FewShotAL, bilayer, fsvid2vid, MarioNETte:AAAI2020}, as they condition image synthesis on 2D facial landmarks. Zakharov \textit{et al.} \cite{Zakharov2019FewShotAL} introduce a few-shot framework that is composed of an identity embedding network and a generator with AdaIN \cite{adain} layers for receiving the embedding vectors. They propose a training strategy consisting of a meta-learning stage which involves optimisation on a multi-person dataset, followed by fine-tuning on an few images of a new and unseen face. The video-based system \cite{fsvid2vid}, namely \textit{few-shot vid2vid}, employs a sequential generator equipped with dynamic SPADE \cite{park2019SPADE} layers, enabling to adapt synthesis on the appearance of new reference images. Zakharov \textit{et al.} \cite{bilayer} propose a SPADE-based system that achieves real-time one-shot talking head synthesis on mobile phones, focusing on the foreground as it disregards background information. In contrast to 2D face modeling, 3D-based approaches such as \textit{Warp-Guided GANs} \cite{Geng2018WarpguidedGF} and \textit{MarioNETte} \cite{MarioNETte:AAAI2020} depend on 3D landmarks to represent faces. As opposed to the preceding works, \textit{MarioNETte} \cite{MarioNETte:AAAI2020} applies a PCA-based disentanglement on identity and expression with 3D facial landmarks, in order to alleviate the identity mismatch problem during reenactment. \textit{HeadGAN} \cite{doukas2020headgan} is among the first one-shot talking head synthesis systems that relies on identity and expression 3DMMs for the disentanglement of expression and identity parameters in dense 3D shapes. \textit{HeadGAN} yields image samples of unprecedented quality in reenactment, while maintaining the identity of the source and exhibiting free-view control.

There is a considerable amount of research involving person-generic \textit{model-free} head synthesis \cite{X2Face, Siarohin_2019_CVPR, FOMM, wang2021facevid2vid}. \textit{X2Face} \cite{X2Face} is one of the pioneering approaches on the problem, which does not rely on any 2D or 3D priors. It is able to animate faces driven by multiple modalities, such as images, audio, and pose codes. In order to address motion transfer for arbitrary objects, Siarohin \textit{et al.} \cite{Siarohin_2019_CVPR} propose \textit{Monkey-Net}, a framework composed of a 2D key-point detector which is jointly trained with a motion prediction and motion transfer network. Key-points are learned directly from data in an unsupervised way, while being assigned to meaningful regions of the objects, in order to enable reliable optical flow approximation for motion transfer. In the follow-up work, \textit{First Order Motion Model for Image Animation (FOMM)}~\cite{FOMM} learns a first order approximation of optical flow, based again on learnable 2D key-points, yielding very promising results. However, during cross-identity motion transfer it uses relative key-points to preserve the identity of the source, which makes the assumption that the face in the first driving frame is in the same pose with the source face. In their recent work, Wang \textit{et al.} \cite{wang2021facevid2vid} apply first order motion approximation on 3D facial key-points. This enables their so called \textit{face-vid2vid} model to perform free-view reenactment, as they can manipulate pose by rotating the 3D points. They further devise two networks, one for predicting canonical key-points and one that estimates pose and expression deformations, which enable to tackle the identity preservation problem. Unlike Wang \textit{et al.} \cite{wang2021facevid2vid}, we do not regress canonical key-points directly, but rather obtain them by removing the estimated head pose and expression deformations from regular 3D key-points, all of which are predicted from images using a unified network.

In Table~\ref{table:background} we present a detailed summary of the key features and design choices of state-of-the-art systems for reenactment and motion transfer, as discussed in this section.

\subsection{Gaze Estimation Systems}

Gaze estimation is the problem of predicting the direction that someone is looking at, based on input images or videos. Ever since~\cite{zhang15_cvpr_mpiigaze} employed CNNs to tackle the task, appearance-based methods have been the default approach. Multiple methods attempt to recover geometric features of eyes and use them to infer gaze~\cite{Park2018ETRA, Park_2018_ECCV_dpge, Wang_2018_CVPR}. Others focus on adapting generic gaze estimation networks to specific test domains, aiming to produce person-specific models~\cite{Park2019ICCV_fewshot, He_2019_ICCVW, Yu_2019_CVPR} or adapting to unseen image domains~\cite{Guo_2020_ACCV, Liu_2021_ICCV}. Recently, significant progress has been made to reduce the labeled data required to build effective gaze estimation systems, by proposing to learn gaze in unsupervised or weakly-supervised settings~\cite{Yu_2020_CVPR, Sun_2021_ICCV, Kothari_2021_CVPR}. In this work, we aim to recover gaze implicitly from the dense geometry of eyes, and employ it for driving our image generator. 

\section{Methodology}\label{sec:Methodology}

Our proposed talking head synthesis system consists of three discrete networks: a network for inferring canonical key-points (Sec. \ref{subsec:canonical}), a gaze estimation network (Sec. \ref{subsec:gaze}) and an image generator (Sec. \ref{subsec:generator}). Each component of \textit{Free-HeadGAN} is trained separately on its individual task.

\subsection{Computation of Canonical Key-points}\label{subsec:canonical}

Given a portrait image, our goal here is to extract a set of key-points in a canonical space, which are independent both from the head pose and facial expressions of the subject. This representation merely depends on the geometry of the input face and therefore encodes only identity-related information. To that end, we propose a neural network $E_{can}$ that learns to a) regress sparse 3D facial key-points $\textbf{p} = \{\textbf{p}_k\}_{k=1,\dots,K}$, with $\textbf{p}_k \in [-1, 1]^3$, b) estimate head pose, as an affine transformation $\mathcal{T}=\{s, \textbf{R}, \textbf{t}\}$ and c) compute a 3D vector $\textbf{d}_k$ for each key-point $k$, that models the non-linearity of deformations caused by facial expressions. The architecture of $E_{can}$ follows that of the head pose and expression deformation estimator in \cite{wang2021facevid2vid}, with the addition of one more affine output layer that predicts $K$ points $\textbf{p}$. 

Canonical key-points are obtained by first estimating the 3D points and then subtracting the expression deformations and removing translation, rotation, and scale:
\begin{equation}
\textbf{p}^{can}_k = \dfrac{1}{s} \textbf{R}^{-1}(\textbf{p}_k - \textbf{d}_k - \textbf{t}), \quad k=1,\dots,K.
\label{eq:1}
\end{equation}
We further define the inverse operation, which brings canonical points back to the original 3D space, by adding scale, rotation, translation as well as the expression deformation:
\begin{equation}
\textbf{p}_k = s \textbf{R}
\textbf{p}^{can}_k + \textbf{t} + \textbf{d}_k, \quad k=1,\dots,K.
\label{eq:2}
\end{equation}

We observed that predicting the expression-related deformations $\textbf{d}_k$ in the original 3D space, as in \cite{wang2021facevid2vid}, yields significantly better results than placing the deformation on the canonical space. 

In order to train network $E_{can}$, we are given a source and a target image pair $(\textbf{x}^{s}, \textbf{x}^{t})$, which depict the same person performing a different and random pose and expression. We use $E_{can}$ to regress 3D key-points as well as pose transformations and expression perturbations from both images: $\textbf{p}^s, \mathcal{T}^s, \textbf{d}^s$ and $\textbf{p}^t, \mathcal{T}^t, \textbf{d}^t$. After that, we apply Eq.~\ref{eq:1} to get the corresponding canonical points $\textbf{p}^{s, can}$ and $\textbf{p}^{t, can}$, as shown in Fig.~\ref{fig:canonical}. Then, based on Eq.~\ref{eq:2} we aim to recover the target key-points, by transforming the canonical representation of the source using the target pose and expression, and vice versa, which results in points $\textbf{p}^{s, rec}$ and $\textbf{p}^{t, rec}$. Having access to image pairs of the same identity is crucial for guiding $E_{can}$ to focus on expression and pose discrepancies between images and thus not include identity-related information in deformation vectors $\textbf{d}^s$ and $\textbf{d}^t$.

We optimise $E_{can}$ in a supervised manner, using $K=68$ 3D facial landmarks estimated by a pre-trained RetinaFace \cite{RetinaFace} model as pseudo-ground truth data. That is, given the ground source and target landmarks $(\textbf{l}^{s}, \textbf{l}^{t})$, we minimise the distance
\begin{equation}
\mathcal{L}^p = ||\textbf{p}^s - \textbf{l}^{s}||_2^2 + ||\textbf{p}^{t} - \textbf{l}^t||_2^2,
\end{equation}
which forces $E_{can}$ to predict accurate facial key-points. We noticed that learning to estimate the 3D key-points helps $E_{can}$ to predict the perturbations caused due to expressions more efficiently. We further minimise the key-point reconstruction distance
\begin{equation}
\mathcal{L}^{rec} = ||\textbf{p}^{s, rec} - \textbf{l}^{s}|||_2^2 + ||\textbf{p}^{t, rec} - \textbf{l}^{t}||_2^2,
\end{equation}
which forces $E_{can}$ to learn the affine pose transformations as well as the deformation vectors, in an effort to recover the source and target key-points. In order to assist $E_{can}$ distinguish between rigid and non-rigid transformations, we penalise the error between the predicted target head rotation $\textbf{R}^t$ and the ground truth $\textbf{R}^{t}_{*}$ (here expressed as Euler angles). Furthermore, a regularisation term on the expression deformation vectors ensures that key-point perturbations due to expressions are kept small, as we want to avoid encoding identity-specific details in these vectors:
\begin{equation}
\begin{split}
\mathcal{L}^R = ||\textbf{R}^t - \textbf{R}^t_*||_2^2, \quad \mathcal{L}^d = ||\textbf{d}^s||_2^2 + ||\textbf{d}^t||_2^2.
\end{split}
\end{equation}
Combining the loss terms defined above, the overall objective function for network $E_{can}$ is given as
\begin{equation}
\mathcal{L}_{E_{can}} = \lambda_p \mathcal{L}^p + \lambda_{rec} \mathcal{L}^{rec} + \lambda_{R} \mathcal{L}^{R} + \lambda_{d}, \mathcal{L}^{d}  
\end{equation}
with the hyper-parameters set to the following values: $\lambda_p = \lambda_{rec} = 200, \lambda_{R}=2$ and $\lambda_{d} = 5$.

\begin{figure}[!t]
\centering
\includegraphics[width=0.49\textwidth]{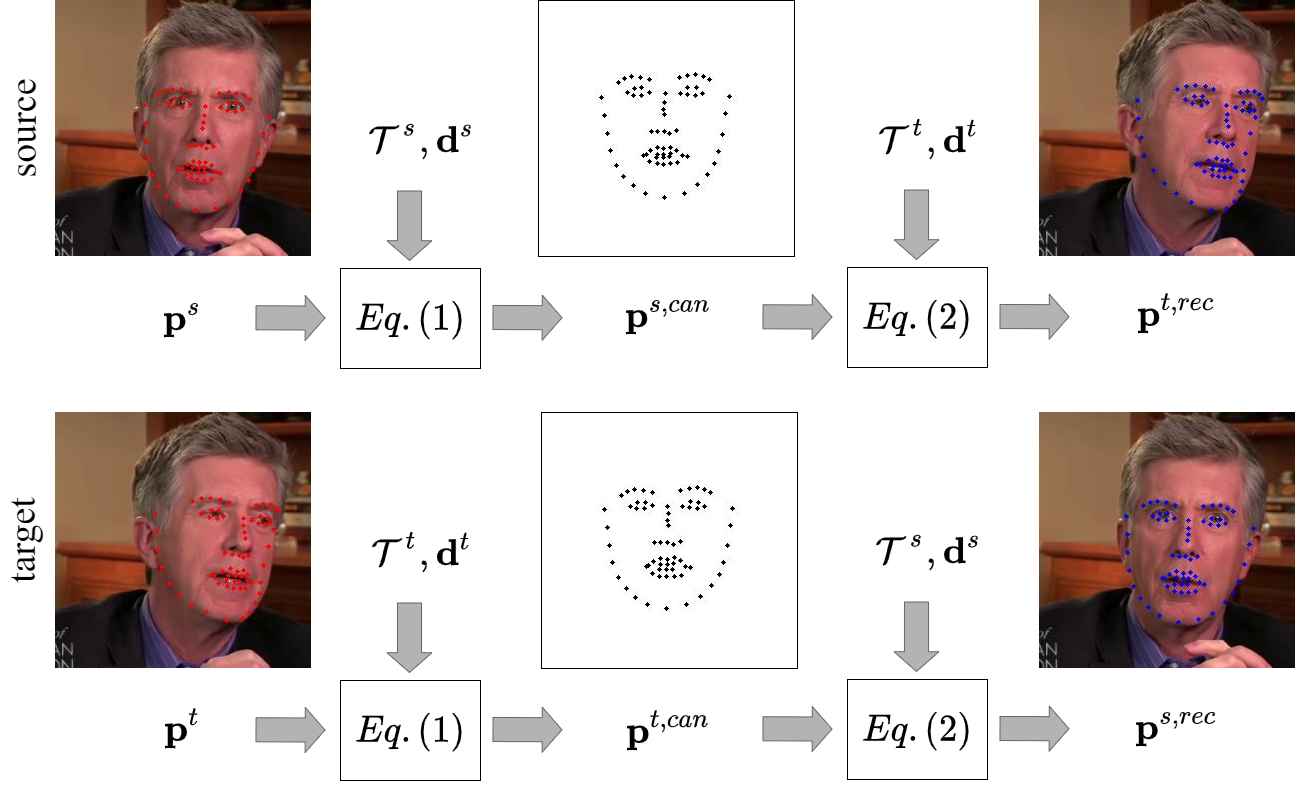}
\caption{Canonical key-points computation pipeline during training. Using a pair of frames from the same person, we train $E_{can}$ to disentangle identity from expression and pose. We project the regressed 3D key-points into a canonical space and then try to reconstruct them, after swapping the canonical points of the source and target images.}
\label{fig:canonical}
\end{figure}

\subsection{Gaze estimation}\label{subsec:gaze}

\begin{figure}[!h]
\centering
\includegraphics[width=0.48\textwidth]{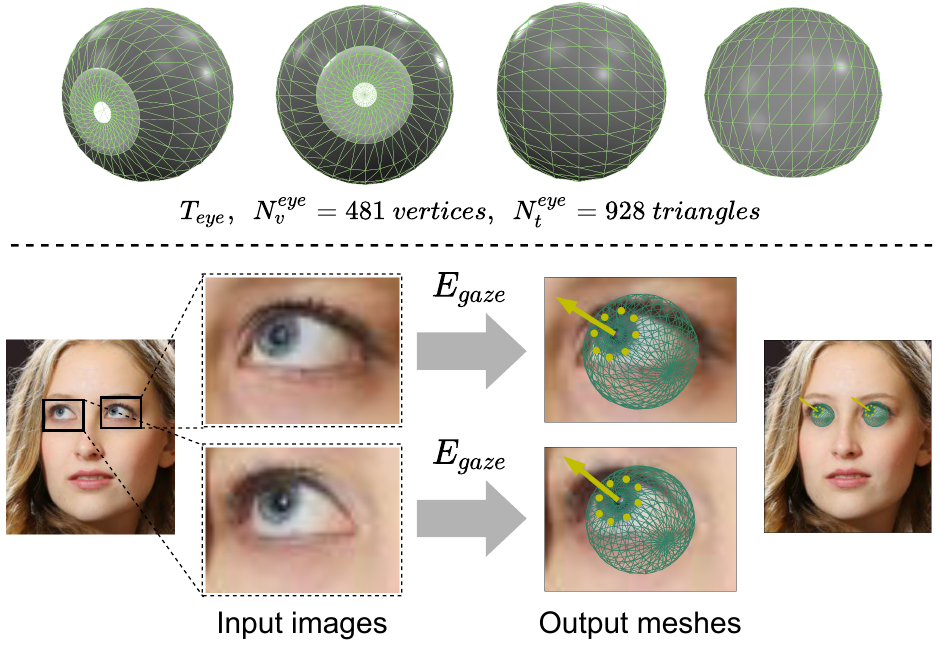}
\caption{Outline of our gaze estimation method. Having defined an eyeball template, $\mathbf{T}_{eye}$ we generate pseudo-ground truth meshes for available gaze estimation dataset and employ them to train our 3D eye mesh regression network $E_{gaze}$, which given input images produces 3D coordinates adhering to $\mathbf{T}_{eye}$.}
\label{fig:Egaze}
\end{figure}

The gaze estimation network $E_{gaze}$ learns to predict the 3D orientation of eyes from an input image of each eye. Particularly, our aim is to recover spherical coordinates $(\theta, \phi)$ that represent gaze direction and employ them for training as well as driving our reenactment system. To that end, we adopt a mesh regression approach to gaze estimation, meaning that we train $E_{gaze}$ to predict 3D meshes of eyes instead of 3D gaze vectors or angles. This approach is based on the fact that estimating dense geometry instead of few pose parameters, has recently been shown to benefit face and body pose estimation systems~\cite{Deng_2020_CVPR, Kulon_2020_CVPR, Guler2018DensePose}. 

To train $E_{gaze}$, we first define a 3D eye mesh template, $\mathbf{T}_{eye}$, consisting of $N_{v}^{eye}=481$ vertices and $N_{t}^{eye}=928$ triangles, as shown in Fig.~\ref{fig:Egaze}, and enforce our model's predictions to adhere to this template so that exact correspondence exists. As common gaze estimation datasets provide gaze annotations as 3D vectors, angles or points on screen~\cite{zhang15_cvpr_mpiigaze, gaze360_2019, cvpr2016_gazecapture}, we create training and validation data compatible with our mesh regression approach by automatically fitting $\mathbf{T}_{eye}$ on available images, based on 2D sparse landmarks around the iris contour and the available gaze labels. To obtain the iris landmarks we employ the network from~\cite{Park2018ETRA}, but any similar model could have been used.   

We adopt a simple architecture for $E_{gaze}$, consisting of a ResNet-34 backbone and a fully connected layer mapping eye image features to  $\mathbf{v}$, which is a vector of $3\times N_{v}^{eye}$ real values representing $N_{v}^{eye}$ 3D eye coordinates on a normalized space. We optimize $E_{gaze}$ in a supervised fashion, based on the 3D eye meshes we fitted on available gaze estimation datasets. To recover 3D points from images, we minimise the distance
\begin{equation}
\mathcal{L}^{v} = ||\mathbf{v} - \mathbf{v}^*||_1,
\label{eq:vertex_loss}
\end{equation}
between predicted coordinates $\mathbf{v}$ and the corresponding pseudo-ground truth ones $\mathbf{v}^*$. Additionally, to maintain feasible eye shape, we minimize the distance 
\begin{equation}
\mathcal{L}^{e} = ||\mathbf{e} - \mathbf{e}^*||_1,
\label{eq:edge_loss}
\end{equation}
between the $3N_{t}^{eye}$ edge lengths $\mathbf{e}$ computed from $\mathbf{v}$ and those, $\mathbf{e}^*$, computed from $\mathbf{v^*}$. We define edge length to be the euclidean distance between vertices of the the same triangle according to $\mathbf{T}_{eye}$. Lastly, to boost gaze estimation accuracy we employ the gaze loss
\begin{equation}
\mathcal{L}^{g} = (180 / \pi) \arccos(\mathbf{g}^T\mathbf{g}^*),
\label{eq:edge_loss}
\end{equation}
where $\mathbf{g}$ and $\mathbf{g}^*$ are normalized 3D gaze vectors, calculated by the centers of the eye and iris, from the predicted and ground truth meshes respectively. Combining the above losses, the overall objective function for $E_{gaze}$ is given as
\begin{equation}
\mathcal{L}_{E_{gaze}} = \lambda_v\mathcal{L}^{v} + \lambda_e\mathcal{L}^{e} + \lambda_g\mathcal{L}^{g},
\label{eq:eyes_loss}
\end{equation}
with the hyper-parameters set to the following values $\lambda_v=\lambda_e=0.1$ and $\lambda_g=1$. As in our reenactment system we code gaze using angles $(\theta, \phi)$, we calculate them from predicted eye meshes using the gaze vector $\mathbf{g}=(g_x, g_y, g_z)$ as
\begin{equation}
\theta = \arctan \sqrt{g_x^2 + g_y^2} / g_z, \ \ \phi = \arctan g_y / g_z.
\label{eq:crt2sph}
\end{equation}

\subsection{Image Synthesis}\label{subsec:generator}

\begin{figure}[!b]
\centering
\includegraphics[width=0.49\textwidth]{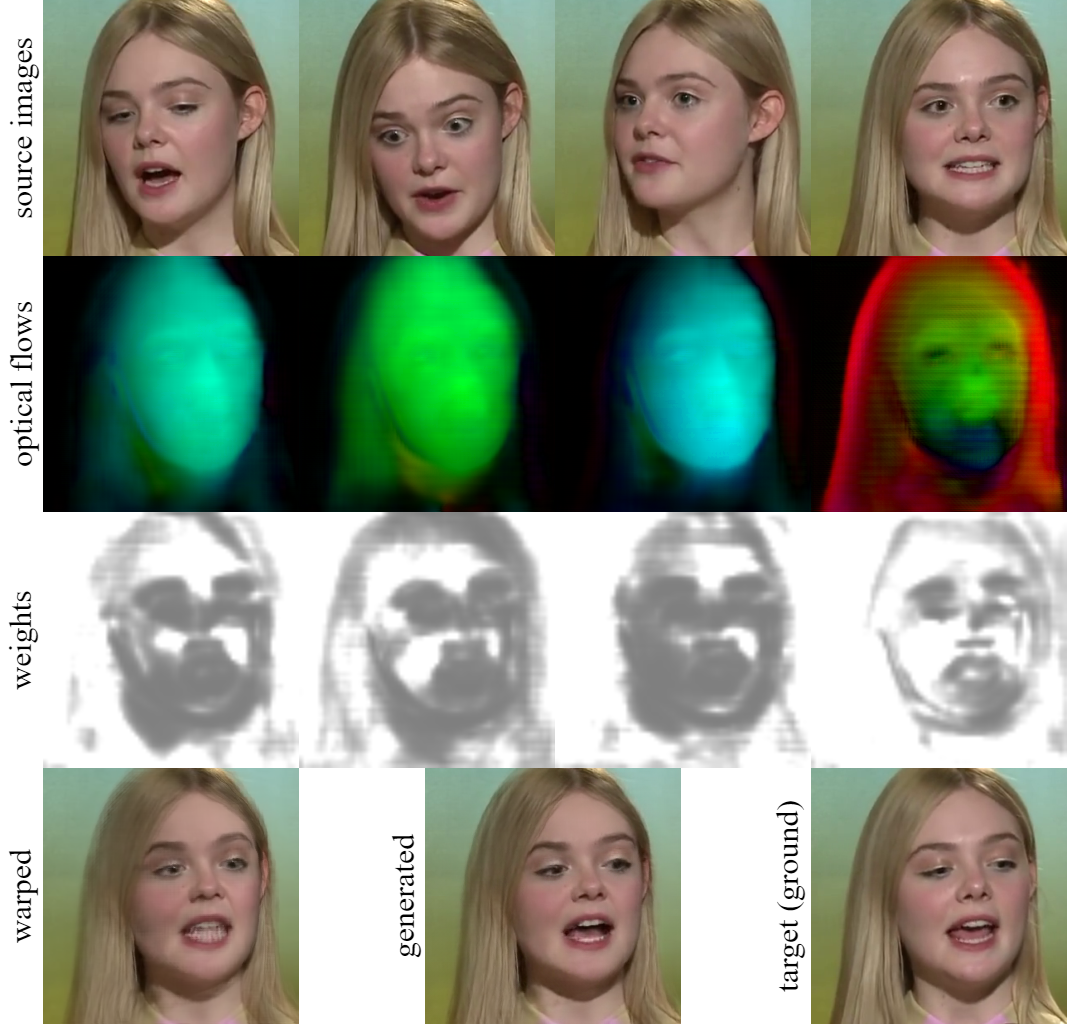}
\caption{Illustration of optical flows and weights predicted by the flow network, in case four source images are provided instead of one (4-shot). The warped image is calculated from the source images, flows and weighs with the application of Eq.~\ref{eq:7}.}
\label{fig:weights}
\end{figure}

\begin{figure*}[!t]
\centering
\includegraphics[width=\textwidth]{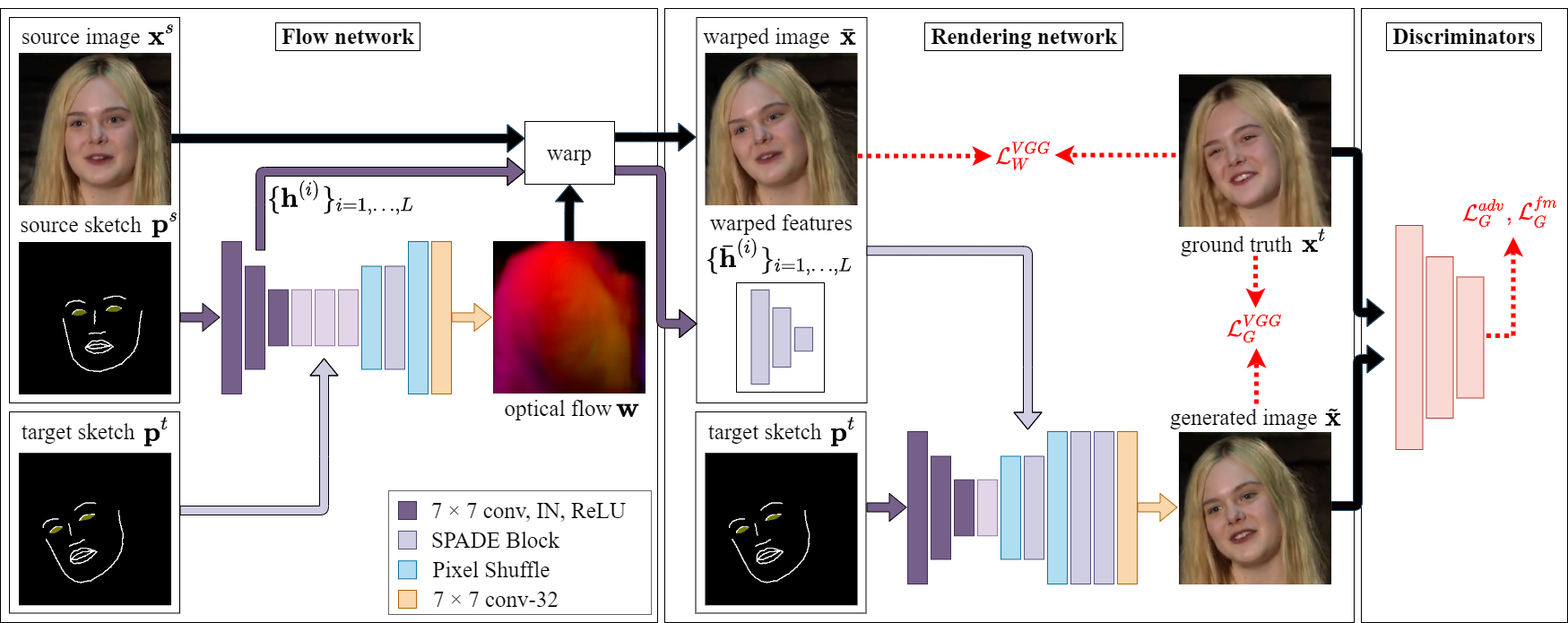}
\caption{Overview of the image generation component. The generator is made up from two modules: a flow network and a rendering network. The flow network computes the optical flow for warping the reference image and features, according to the target sketch. The rendering network uses this visual information, in order to translate the target sketch into a photo-realistic image of the source. The two networks are trained jointly in an adversarial manner, with the assistance of image discriminators.}
\label{fig:facesynthesis}
\end{figure*}

We perform talking head synthesis with the assistance of an image-to-image translation network $G$, which is based on the generator of HeadGAN \cite{doukas2020headgan}, without AdaIN layers\cite{adain} as we do not processing audio features. Given a source image $\textbf{x}^s$ and a target frame $\textbf{x}^t$, along with the corresponding facial key-points $\textbf{p}^s$, $\textbf{p}^t$ and gaze angles $\textbf{g}^s$, $\textbf{g}^t$, we draw a source and a target image sketch. Gaze angles are color coded in the sketches, within the areas defined by key-points that belong to the eyes. The two sketches, together with the source image, serve as inputs to network $G$, which learns to generate a photo-realistic image $\tilde{\textbf{x}}$ of the source, in the target head pose and facial expression. In more detail, as in \cite{doukas2020headgan} network G is comprised of two modules: a flow network and a rendering network. 

First, the flow network extracts visual feature maps from the source image and its corresponding key-point sketch in multiple spatial resolutions: $\{\textbf{h}^{(i)}\}_{i=1,\dots,L}$, $L=3$. Then, the target sketch is injected into the network through SPADE \cite{park2019SPADE} layers, as modulation input, and guides the prediction of the optical flow $\textbf{w}$, which warps the source image to the target expression and pose. In addition, we perform warping on each feature map, in order to align them spatially with the desired expression and pose. In practice, we utilise the backward optical flow warping operator from \textit{FlowNet 2.0} \cite{flownet2-pytorch, IMKDB17}.

The rendering network passes the target sketch through an encoder and combines the extracted feature map with the warped features $\{\bar{\textbf{h}}^{(i)} = \textbf{w}(\textbf{h}^{(i)}) \}_{i=1, \dots, L}$ and warped image $\bar{\textbf{x}} = \textbf{w}(\textbf{x}^s)$, which enter the network through SPADE \cite{park2019SPADE} layers, as modulation inputs. Up-sampling is performed with PixelShuffle \cite{pixelshuffle} layers. The final output is a photo-realistic image $\tilde{\textbf{x}}$ of the source identity, imitating the facial expressions and head pose shown in the target image.

\noindent \textbf{N-shot extension.} We further extend our method to enable few-shot learning, in cases where more than one source images are available. For that, we propose an optional attention mechanism, with the addition of one more output layer to the flow network, which now learns to compute a set of 2D weights $\textbf{m} \in \mathbb{R}^{H \times W}$, alongside the optical flow field. Given $M$ source frames $\{\textbf{x}^s_j\}_{j=1,\dots,M}$, we pass each one through the flow network, gaining flows $\{\textbf{w}_j\}_{j=1,\dots,M}$ and weights $\{\textbf{m}_j\}_{j=1,\dots,M}$. Next, the warped image is computed with the assistance of a Softmax function
\begin{equation}
    \bar{\textbf{x}} = \dfrac{\sum_j^M \exp(\textbf{m}_j) \textbf{w}_j(\textbf{x}^s_j)}{\sum_j^M \exp(\textbf{m}_j)}
    \label{eq:7}
\end{equation}
and the warped features are given as
\begin{equation}
    \bar{\textbf{h}}^{(i)} = \dfrac{\sum_j^M \exp(\textbf{m}_j) \textbf{w}_j(\textbf{h}^{(i)}_j)}{\sum_j^M \exp(\textbf{m}_j)}, \quad i=1,\dots,L.
\end{equation}
Fig. \ref{fig:weights} shows a visual example of flows and weights.

\begin{figure*}[!ht]
\centering
\includegraphics[width=\textwidth]{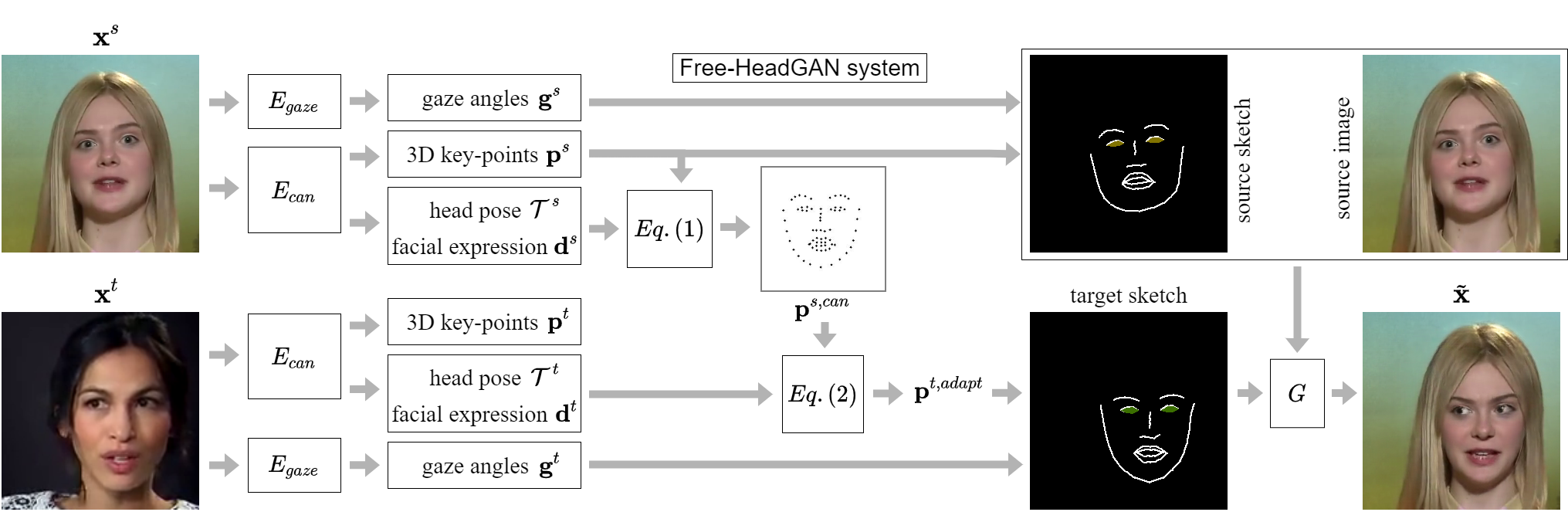}
\caption{End-to-end pipeline of \textit{Free-HeadGAN} during inference, for the task of reenactment. We adapt the target key-points to the facial shape (identity) of the source.}
\label{fig:system}
\end{figure*}

\noindent \textbf{Training.} The optical flow and rendering networks that make $G$ are trained jointly, on the task of self-reenactment (reconstruction). To that end, we sample the source image from the frames of the target video to obtain image pairs that belong to the same person and scene. In this case, the generated image $\tilde{\textbf{x}}$ should match the target frame $\textbf{x}^t$ that serves as ground truth. We optimise the generator G by minimising the distance between feature maps extracted from multiple layers of a pre-trained VGG network \cite{vggloss}:
\begin{equation}
   \mathcal{L}_G^{VGG} = \sum_l || VGG_l(\tilde{\textbf{x}}) - VGG_l(\textbf{x}^t) ||_1.
\end{equation}
Additionally, we apply a VGG loss on the warped image, in order to force the flow network to learn a correct flow from the source image to the desired head pose, which gives the loss term $\mathcal{L}_F^{VGG}$. We further improve the photo-realism of synthetic images by placing an adversarial loss term $\mathcal{L}_G^{adv}$ on $\tilde{\textbf{x}}$ and more specifically a Hinge GAN loss \cite{lim2017geometric}. Similarly with \cite{head2head2020} and \cite{doukas2020headgan}, we employ two critics, a general image discriminator $D_I$ and a dedicated discriminator $D_M$ for the mouth area, which are optimised alongside G. Finally, we use these discriminators to compute visual features from both real (target) and synthetic images and compute a feature matching loss $\mathcal{L}_G^{fm}$ \cite{xu2017learning} that is known to be effective at increasing the photo-realism of synthetic samples. Summing up, the overall objective function for G is given by:
\begin{equation}
\begin{split}
   \mathcal{L}_G = & \mathcal{L}_G^{adv} + \lambda_{VGG} (\mathcal{L}_G^{VGG} + \mathcal{L}_F^{VGG}) + \lambda_{FM} \mathcal{L}_G^{FM}
\end{split}
\end{equation}
where $\lambda_{VGG} = \lambda_{FM} = 10$. Both discriminators are optimised under their corresponding adversarial loss terms and have a similar architecture with the one proposed in \cite{park2019SPADE}. Please refer to Fig.~\ref{fig:facesynthesis} for an illustration of our model.  

\subsection{Free-HeadGAN inference}

During inference, in the task of reenactment the source and target images belong to different identities. In order to adapt the target key-points to the facial shape of the source identity, we use $E_{can}$ to regress 3D key-points, head pose and expression deformations from $\textbf{x}^{s}$, and then evaluate Eq.~\ref{eq:1} to obtain the canonical key-points $\textbf{p}^{s, can}$. At the same time, we estimate the target pose and expression $\mathcal{T}^t, \textbf{d}^t$ from $\textbf{x}^{t}$. The adapted target key-points $\textbf{p}^{s, adapt}$ are obtained with the application of Eq.~\ref{eq:2}, which transforms the source canonical key-points using the estimated target pose and expression. In this way, we remove the target identity-related information from key-points and inject the source one, which helps to overcome to problem of identity mismatching. In parallel, we estimate eye gaze from both images with $E_{gaze}$ network and finally draw the sketches that serve as input to generator $G$, which hallucinates the output image $\tilde{\textbf{x}}$. For a visual inspection of our proposed pipeline in reenactment please refer to Fig~\ref{fig:system}.

\section{Experiments}

\subsection{Dataset and Training}
We train each component of \textit{Free-HeadGAN} independently. Both the network that infers the canonical key-points $E_{can}$ and image synthesis network $G$ (with discriminators $D_I$, $D_M$) are optimised on VoxCeleb \cite{voxceleb} video dataset, which contains over one hundred thousand clips and more than one thousand identities, at a 256 $\times$ 256 resolution. We keep the original train and test split. As a pre-processing step, we prepare three-dimensional 68 landmarks for each frame in the training split with RetinaFace model \cite{RetinaFace}, which serve as pseudo-ground truth annotations. We note that RetinaFace is pre-trained on WIDERFACE dataset \cite{yang2016wider}. For the extraction of ground truth head rotation angles, we employ a least-squares solver to determine the transformation between facial landmarks and a fixed landmark template that represents the frontal-neutral pose. The aforementioned annotations are used only for training \textit{Free-HeadGAN}. During inference, our method requires only the pair of a source and a target image.

For the optimisation of network $E_{can}$, we use ADAM solver \cite{adam} with $\beta_1 = 0.5$, $\beta_2 = 0.999$ and learning rate $\eta = 0.0002$. We use the same optimiser and hyper-parameters for training $G$ and its adversaries $D_I$, $D_M$. All models are optimised for 5 epochs on the entire VoxCeleb database. We extend \textit{Free-HeadGAN} to accommodate N-shot learning, by first training in one-shot without predicting attention weights. Then, we add the weight layer and fine-tune the GAN for one more epoch using two source images (2-shot), while freezing the parameters of the flow network. 

We optimize network $E_{gaze}$ on a combination of recent gaze estimation datasets, aiming to include variation from multiple image domains. Particularly we employ ETH-XGaze~\cite{Zhang2020ETHXGaze} which includes large variation in gaze and face pose and consists of 756 thousand frames from 80 subject, Gaze360~\cite{gaze360_2019} which is captured both indoors and outdoors and includes 127 thousand training sequences from 365 subjects and MPIIGaze~\cite{zhang15_cvpr_mpiigaze} which provides over 213 thousand frames of 15 subjects, captured with laptop cameras. We employ the default training and validation sets from Gaze360 and MPIIGaze, while we perform a manual split for ETH-XGaze as gaze labels for its test data are not available. Lastly, we use ADAM optimizer \cite{adam} with $\beta_1 = 0.9$, $\beta_2 = 0.999$ and learning rate $\eta = 0.0001$.

\begin{figure*}[!ht]
\centering
\subfloat[Self-reenactment example]{\includegraphics[scale=0.26]{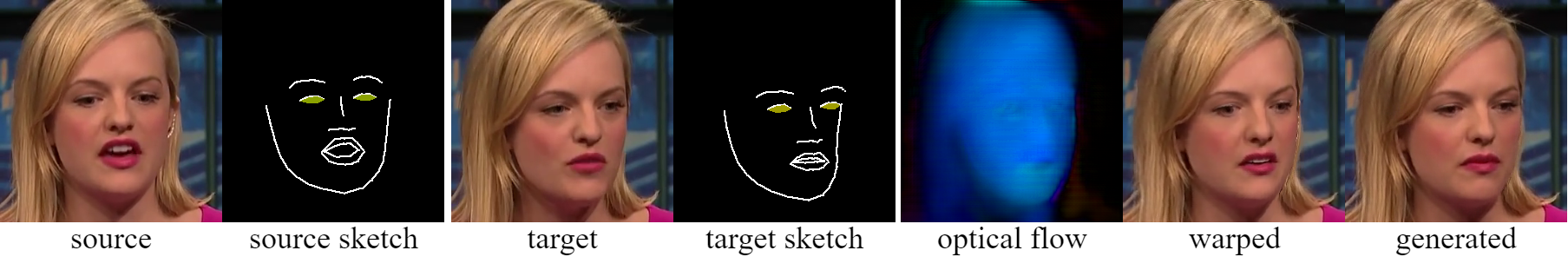}
\label{fig:reconstruction}} \\
\subfloat[Reenactment example]{\includegraphics[scale=0.26]{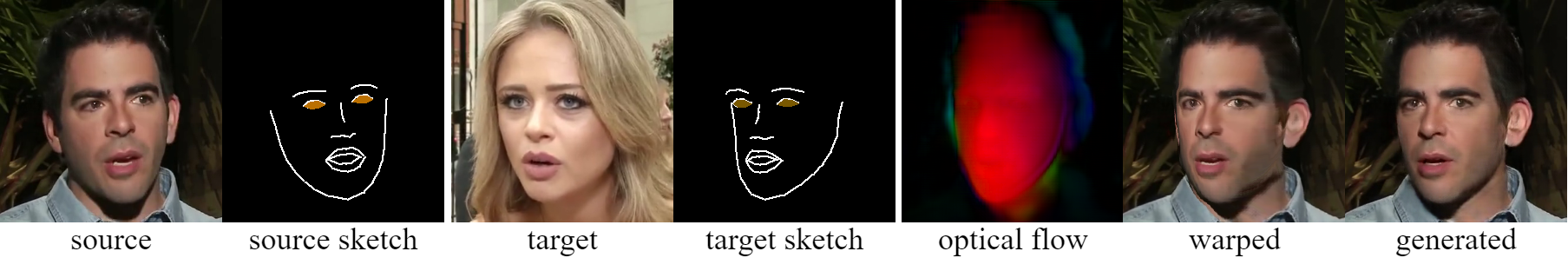}
\label{fig:reconstruction}}
\caption{We evaluate the generative performance of \textit{Free-HeadGAN} on the tasks of (a) same-identity reconstruction (self-reenactment) and (b) cross-identity motion transfer (reenactment).}
\label{fig:reconstruction_reenactment}
\end{figure*}

\begin{table*}[!hb]
\renewcommand{\arraystretch}{1.15}
\caption{Numerical comparison with state-of-the-art methods on the tasks of self-reenactment (same-identity reconstruction) and reenactment (cross-identity motion transfer) for VoxCeleb \cite{voxceleb} test set.}
\label{table:comparison}
\begin{center}
\begin{tabular}{|c||c|c|c|c|c|c||c|c|c|c|c|}
\hline
 & \multicolumn{6}{c||}{Self-reenactment} & \multicolumn{5}{c|}{Reenactment} \\
\hline
Method & \textbf{L1} $\downarrow$ & \textbf{PSNR} $\uparrow$ & \textbf{LPIPS} $\downarrow$ & \textbf{FID} $\downarrow$ & \textbf{FVD} $\downarrow$ & \textbf{CSIM} $\uparrow$ & \textbf{FID} $\downarrow$ & \textbf{CSIM} $\uparrow$ & \textbf{ARD} $\downarrow$ & \textbf{AU-H} $\downarrow$ & \textbf{AGD} $\downarrow$ \\
\hline\hline
\textit{X2Face}~\cite{X2Face} & 13.49 & 20.69 & 0.260 & 130.2 & 697 & 0.600 & 122.1 & 0.520 & 4.39 & 0.346 & 21.4\\
\textit{fs-vid2vid}~\cite{fsvid2vid} & 17.15 & 18.52 & 0.197 & 62.8 & 471 & 0.542 & - & - & - & - & - \\
\textit{Bi-layer}*~\cite{bilayer} & 12.18 & 20.19 & 0.152 & 92.2 & 394 & 0.590 & 172.8 & 0.563 & \textbf{1.01} & \textbf{0.296} & 16.6 \\
\textit{FOMM-abs}~\cite{FOMM} & 12.34 & 20.93 & 0.153 & 64.9 & 338 & 0.754 & 100.7 & 0.587 & 1.46 & 0.298 & 13.8 \\
\textit{FOMM-rel}~\cite{FOMM} & - & - & - & - & - & - & 63.7 & 0.765 & 12.53 & 0.400 & 21.2 \\
\textit{HeadGAN}~\cite{doukas2020headgan} & 11.32 & 21.46 & 0.112 & 36.1 & 254 & 0.807 & 58.0 & 0.688 & 1.35 & 0.326 & 16.8 \\
\textit{Free-HeadGAN} & \textbf{9.96} & \textbf{22.16} & \textbf{0.100} & \textbf{35.4} & \textbf{248} & \textbf{0.810} & \textbf{53.9} & \textbf{0.789} & 1.26 & 0.351 & \textbf{13.3} \\
\hline
\end{tabular}
\end{center}
\end{table*}

\subsection{Evaluation Metrics}
 
We evaluate the reconstruction capabilities of our method quantitatively, using L1 distance (L1), Peak signal-to-noise ratio (PSNR) and Learned Perceptual Image Patch Similarity (LPIPS) \cite{lpips}. All three metrics measure the distance, between the synthesised and ground truth target frames. L1 distance is computed across RGB channels that are in the range [0, 255]. PSNR is the ratio between the maximum possible power of a signal and the power of noise that affects its correctness. It is defined as $20 \cdot \log_{10} (\mathit{MAX}_I) - 10 \cdot \log_{10} \mathit{MSE}$, where $\mathit{MAX}_I=255$ and $\mathit{MSE}$ denotes the mean squared distance. LPIPS \cite{lpips} is another widely used metric to measure the fidelity of reconstruction, which uses a pre-trained AlexNet model for the extraction of a feature maps. The similarity score between two images is calculated as the distance of their visual features. 

Furthermore, we assess the photo-realism of generated images with Fréchet Inception Distance (FID) \cite{NIPS2017_8a1d6947, Seitzer2020FID} and Fréchet Video Distance (FVD)~\cite{unterthiner2018towards}, as we handle video data and therefore crucial to measure the performance of models considering the temporal coherence among frames. 

The identity preservation in the synthesised data is calculated with Cosine Similarity (CSIM) between embedding vectors from the target and corresponding generated images. All embeddings are computed with the assistance of ArcFace \cite{deng2018arcface}. For reenactment, where we have no access to ground truth data, we extract the embedding from the source image(s) and compare it with the embeddings coming from generated data. This leads to lower CSIM values, as the source and target pose do not usually match and ArcFace's output is slightly affected by poses. 

We use Action Units Hamming distance (AU-H) to measure expression transferability of models. OpenFace \cite{openface} with \cite{AUs}, runs on target and synthetic images for the detection of Action Units (AU). OpenFace recognises whether or not a set of AUs is present in a facial image, with the prediction of an AU boolean vector. We measure the Hamming distance between boolean vectors that are extracted from the corresponding target and synthetic data. 

We measure the correctness of pose transfer with Average Rotation Distance (ARD), which is the $l1$-distance of Euler angles that correspond to head pose, between the target and generated frames in degrees. 

Finally, we evaluate gaze transfer with Average Gaze Distance (AGD). Given the gaze vector $\textbf{g}^t$ regressed with $E_{gaze}$ module from the target frame, and the vector $\tilde{\textbf{g}}$ extracted from the corresponding synthetic image, we compute the angle between vectors as $(180 / \pi) \arccos{(\tilde{\textbf{g}}^{\top} \textbf{g}^t)}$ and average across frames to obtain AGD in degrees.

\subsection{Comparison with State-of-the-Art} \label{sec:comparison}

We compare our one-shot \textit{Free-HeadGAN} system both numerically and visually with state-of-the-art methods under two setups: a) same-identity reconstruction or self-reenactment, where the source and target identities coincide, and b) cross-identity motion transfer or reenactment, in which source and target identities are different. Please refer to Fig.~\ref{fig:reconstruction_reenactment} for a visualisation of the two tasks. More specifically, we compare our approach with \textit{X2Face}~\cite{X2Face}, \textit{few-shot vid2vid (fs-vid2vid)}~\cite{fsvid2vid}, \textit{Bi-layer Neural Avatars (Bi-layer)}~\cite{bilayer}, \textit{First Order Motion Model (FOMM)}~\cite{FOMM} and \textit{HeadGAN}~\cite{doukas2020headgan}. Please note that we tested two variations of \textit{FOMM}, one with absolute (\textit{FOMM-abs}) and one with relative key-point coordinates (\textit{FOMM-rel}), which is the default setting for cross-identity motion transfer. For \textit{X2Face}~\cite{X2Face}, \textit{FOMM}~\cite{FOMM} and \textit{HeadGAN}~\cite{doukas2020headgan}, we use the models provided by their authors, all trained on VoxCeleb~\cite{voxceleb} dataset. For \textit{Bi-layer} method, we used the network parameters provided by its authors, trained on VoxCeleb2 dataset* \cite{Chung18b}. We trained \textit{fs-vid2vid}~\cite{fsvid2vid} using the official open source implementation, as pre-trained models are available. Given that the source code for \textit{MarioNETte} \cite{MarioNETte:AAAI2020}, \textit{Warp-guided GANs}~\cite{Geng2018WarpguidedGF} and \textit{face-vid2vid}~\cite{wang2021facevid2vid} is not publicly available, we were not able to measure \textit{Free-HeadGAN}'s performance against these systems.
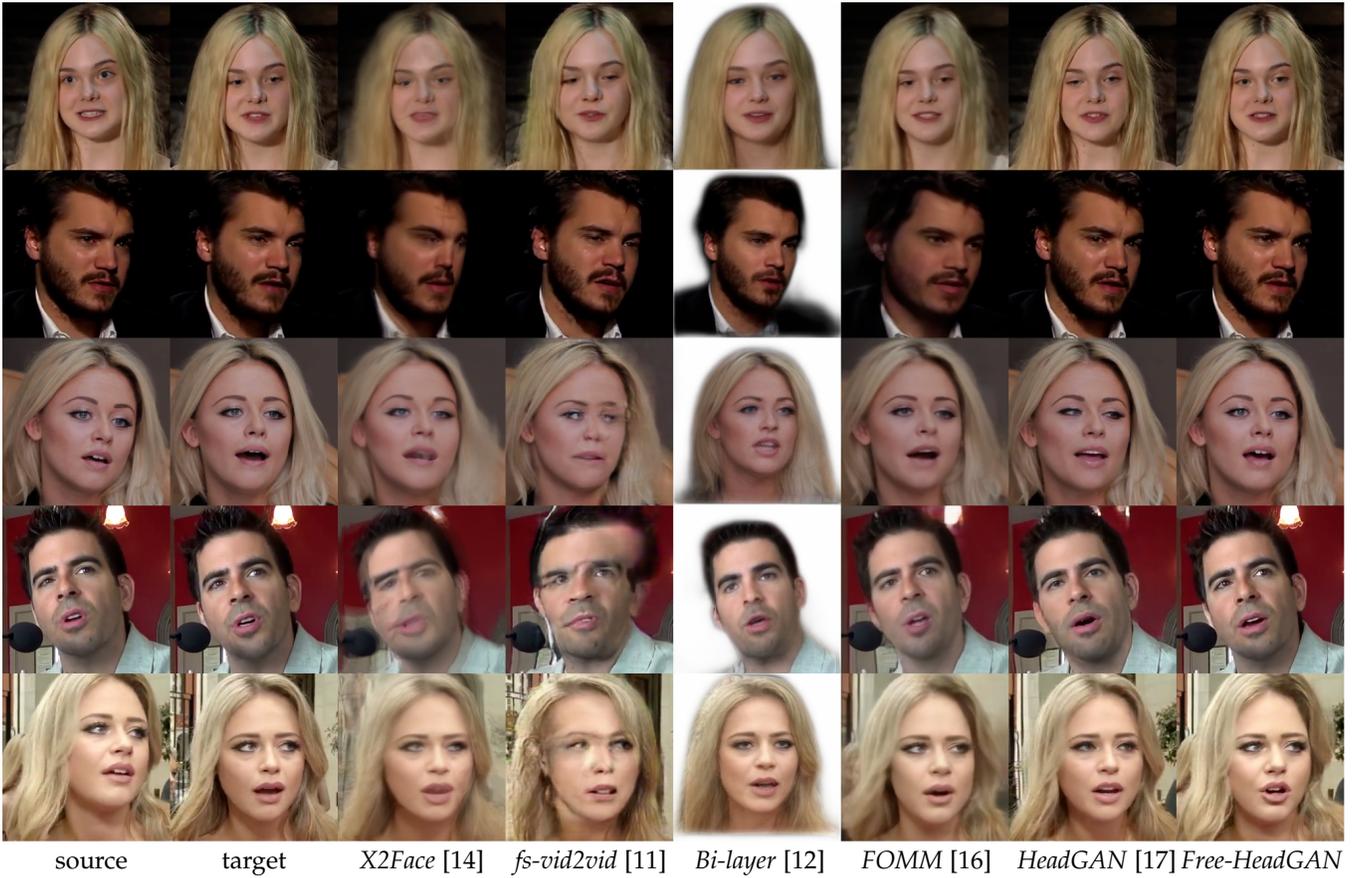
\begin{figure*}[!ht]
\centering
\begin{picture}(520,330)
\put(0,10){\includegraphics[width=0.98\textwidth]{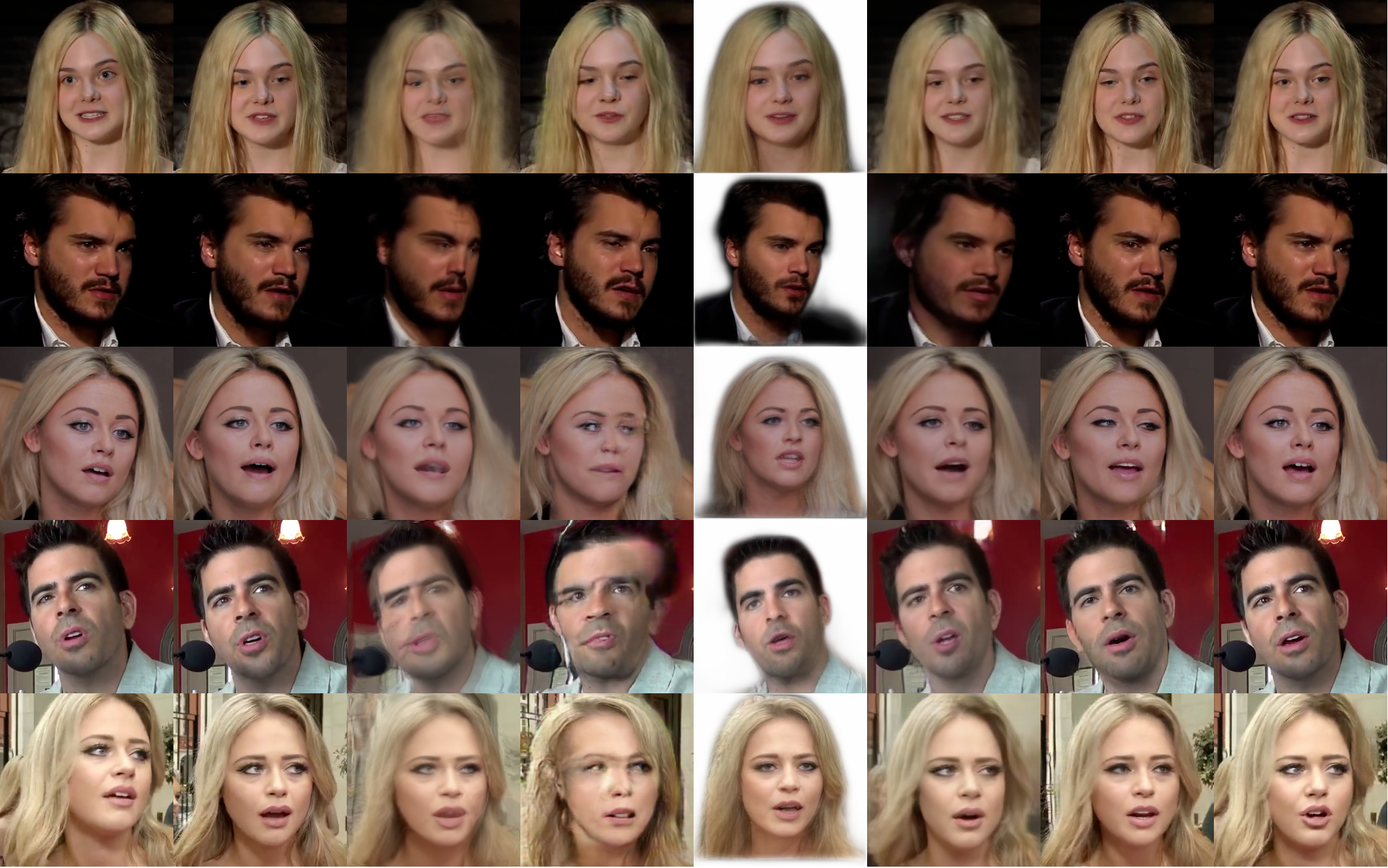}}
\put(20,0){source}
\put(83,0){target}
\put(135,0){\textit{X2Face}~\cite{X2Face}}
\put(194,0){\textit{fs-vid2vid}~\cite{fsvid2vid}}
\put(262,0){\textit{Bi-layer}~\cite{bilayer}}
\put(325,0){\textit{FOMM}~\cite{FOMM}}
\put(384,0){\textit{HeadGAN}~\cite{doukas2020headgan}}
\put(446,0){\textit{Free-HeadGAN}}
\end{picture}
\caption{Visual comparison of our method with baselines on the task of self-reenactment.}
\label{fig:comparison_rec}
\end{figure*}
\begin{figure*}[!hb]
\centering
\begin{picture}(520,300)
\put(0,10){\includegraphics[width=0.98\textwidth]{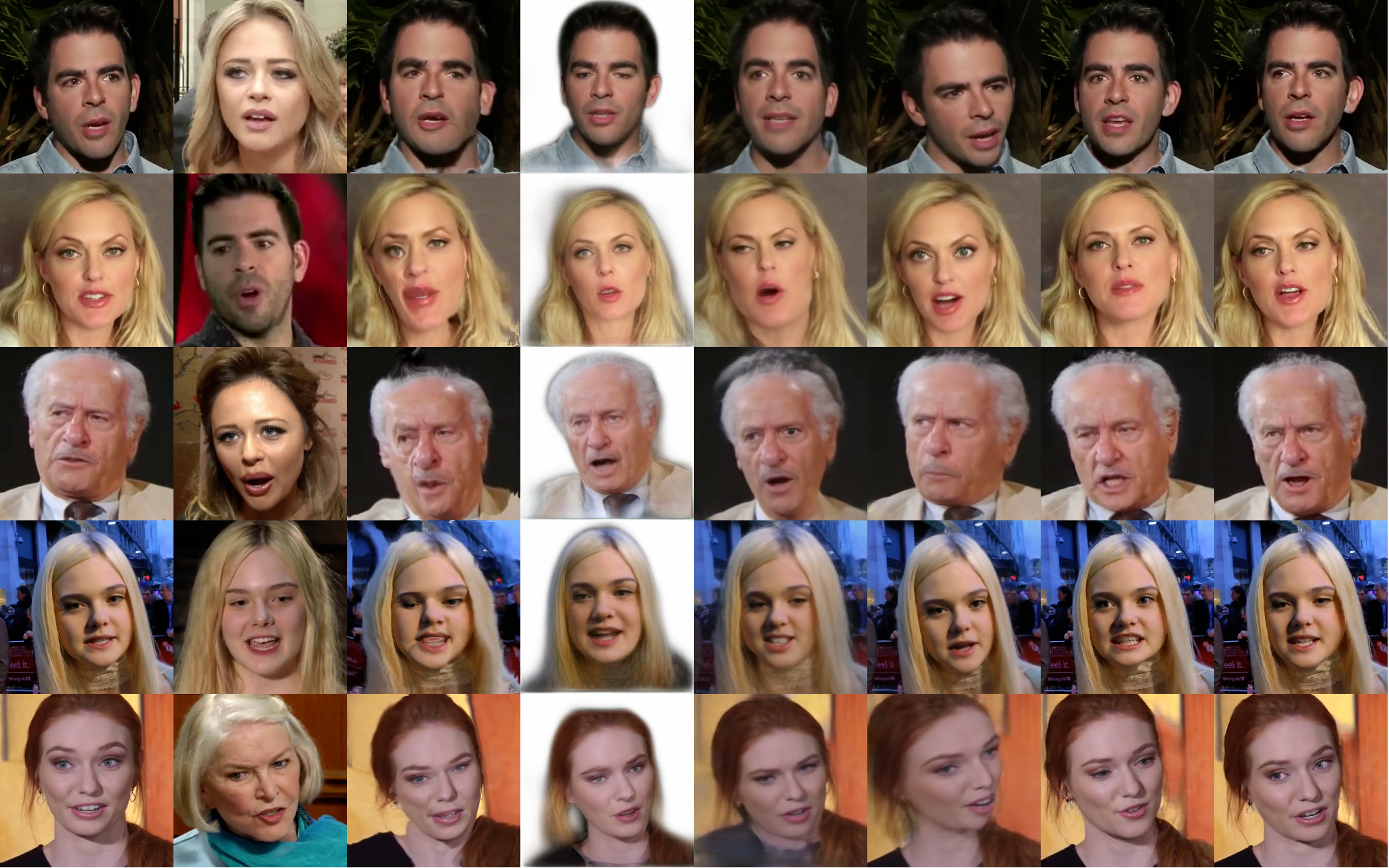}}
\put(20,0){source}
\put(83,0){target}
\put(135,0){\textit{X2Face}~\cite{X2Face}}
\put(198,0){\textit{Bi-layer}~\cite{bilayer}}
\put(252,0){\textit{FOMM-abs}~\cite{FOMM}}
\put(319,0){\textit{FOMM-rel}~\cite{FOMM}}
\put(384,0){\textit{HeadGAN}~\cite{doukas2020headgan}}
\put(446,0){\textit{Free-HeadGAN}}
\end{picture}
\caption{Qualitative evaluation of our method against baselines on the task of reenactment.}
\label{fig:comparison_reen}
\end{figure*}
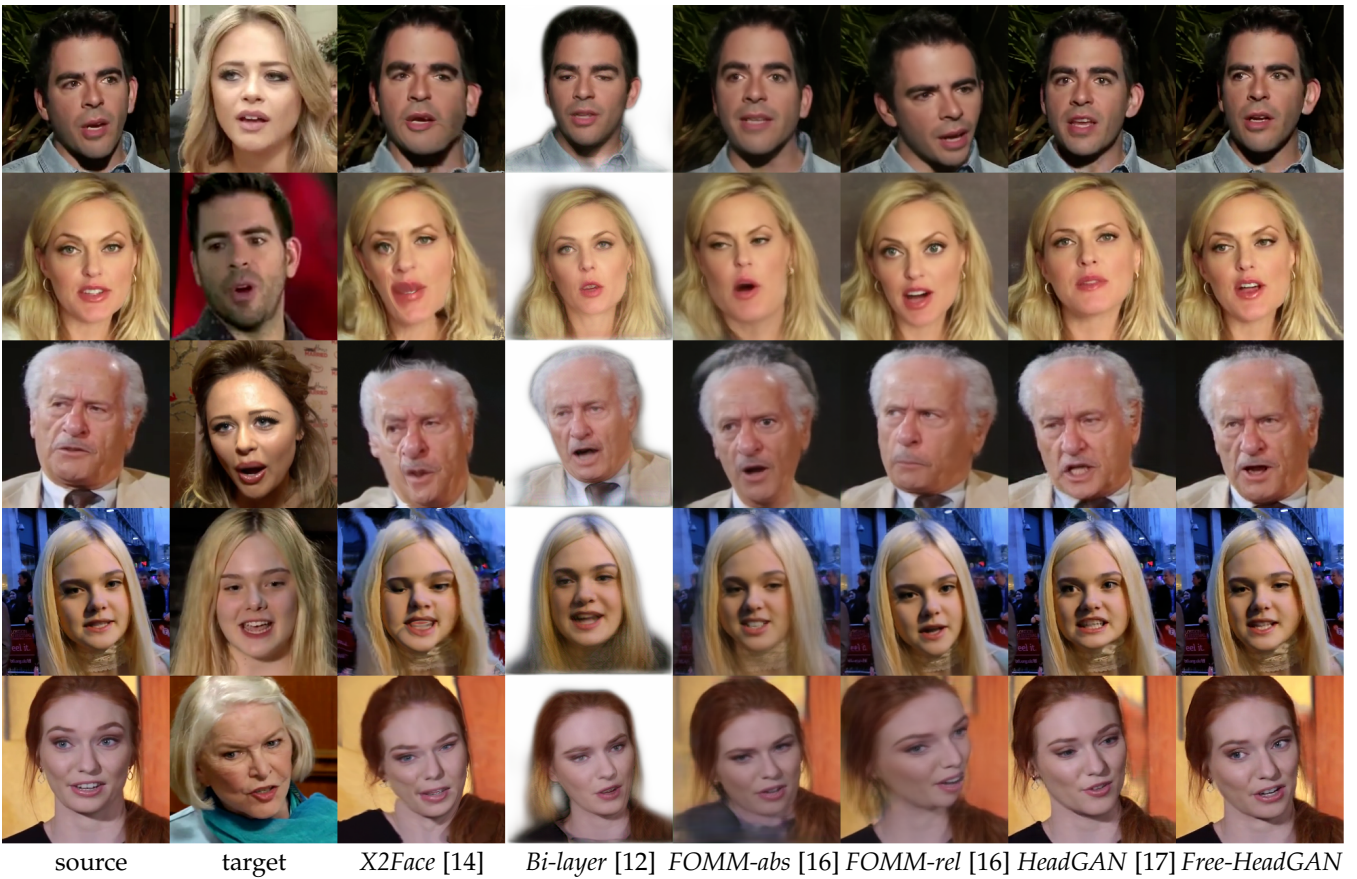

In Table~\ref{table:comparison}, we present a quantitative comparison of the proposed \textit{Free-HeadGAN} with the aforementioned baselines. For self-reenactment, we reconstruct the entire test set of VoxCeleb, while we have chosen 15 pairs of reference images and driving videos for reenactment, the same as in \cite{doukas2020headgan}. As suggested by the results, our method creates superior samples, with respect to image quality and reconstruction fidelity, both in same-identity reconstruction and cross-identity motion transfer. In addition, our approach  preserves the identity of the source better in all experiments. For expression transferability, we observe that \textit{Free-HeadGAN} is left slightly behind from \cite{bilayer} and \cite{doukas2020headgan}. This is attributed to the following two reasons: First, the ground truth 3D facial landmarks annotations have been extracted with \textit{RetinaFace} \cite{RetinaFace}, which is not a model focusing exclusively on landmark prediction and occasionally misses fine details. Second, during reenactment the adaptation of target key-points to the source identity with network $E_{can}$ relies on the regression of expression deformations, which also comes with small inaccuracies. This is compromise we make, as we have chosen 3D facial key-points instead of 2D and we pay particular attention to identity preservation of the source with network $E_{can}$.

In Fig.~\ref{fig:comparison_rec} and Fig.~\ref{fig:comparison_reen} we show examples from a visual comparison of \textit{Free-HeadGAN} with the baselines, on the task of reconstruction and reenactment respectively. As can be seen, the qualitative results confirm our numerical analysis. More specifically, our proposed system outperforms all previous state-of-the-art methods in terms of reconstruction fidelity, photo-realism, identity conservation and gaze transfer. The generated samples of \textit{HeadGAN} \cite{doukas2020headgan} are comparable to ours in terms of visual quality. Nonetheless, our method appears to be significantly more reliable on eye gaze transfer than \textit{HeadGAN}, thanks to our gaze prediction network and explicit gaze conditioning for image synthesis. We urge the reader to refer to our supplementary video for a better inspection of our synthesised data.

\subsection{Ablation Study}

\begin{table}[!b]
\renewcommand{\arraystretch}{1.15}
\caption{Evaluation of \textit{Free-HeadGAN} N-shot extension, in case more than one source images are available.}
\label{table:N-shot}
\centering
\begin{tabular}{|c||c|c|c|c|c|c|}
\hline
N-shot & \textbf{L1} $\downarrow$ & \textbf{PSNR} $\uparrow$ & \textbf{LPIPS} $\downarrow$ & \textbf{FID} $\downarrow$ & \textbf{FVD} $\downarrow$ & \textbf{CSIM} $\uparrow$ \\
\hline
1-shot & 9.96 & 22.16 & 0.100 & 35.4 & 248 & 0.810 \\
2-shot & 9.02 & 22.99 & 0.088 & 32.7 & 224 & 0.834 \\
4-shot & 7.90 & 24.01 & 0.077 & 30.7 & 207 & 0.857 \\
8-shot & 7.20 & 24.96 & 0.068 & 29.9 & 196 & 0.868 \\
\hline
\end{tabular}
\end{table}

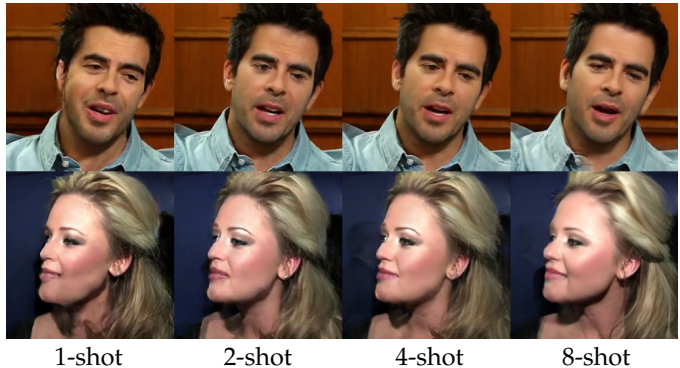
\begin{figure}[!h]
\centering
\begin{picture}(260,140)
\put(0,10){\includegraphics[width=0.49\textwidth]{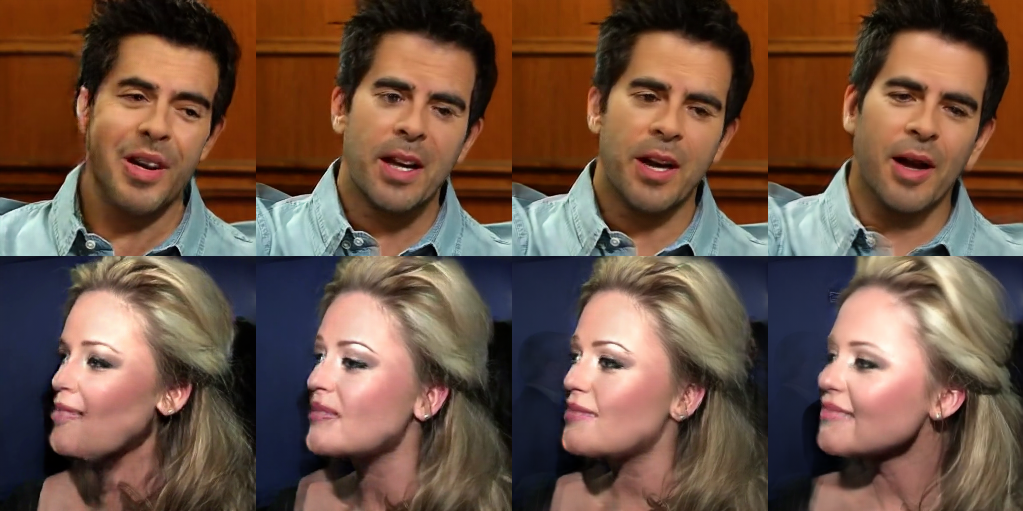}}
\put(18,0){1-shot}
\put(82,0){2-shot}
\put(147,0){4-shot}
\put(210,0){8-shot}
\end{picture}
\caption{Synthetic examples when using one, two, four or eight source images in the N-shot scenario. The results improve in terms of image quality as we get access to more source inputs.}
\label{fig:shot}
\end{figure}

\begin{table}[!b]
\renewcommand{\arraystretch}{1.15}
\caption{Ablation study results on the significance of \textit{Free-HeadGAN} components on the task of reenactment.}
\label{table:ablation}
\centering
\begin{tabular}{|c||c|c|c|c|}
\hline
Variation & \textbf{FID} $\downarrow$ & \textbf{CSIM} $\uparrow$ & \textbf{AU-H} $\downarrow$ & \textbf{AGD} $\downarrow$ \\
\hline
\textit{Free-HeadGAN} & \textbf{53.9} & \textbf{0.789} & 0.351 & \textbf{13.3} \\
\textit{Free-HeadGAN} w/o $E_{can}$ & 66.6 & 0.640 & 0.304 & - \\
\textit{Free-HeadGAN} w/o $E_{gaze}$ & - & - & - & 17.7 \\
\hline
\end{tabular}
\end{table}

We follow a few-shot learning approach to adjust our system in a setting where multiple source images are available. Although our flow network extended to predict weights has been trained in a 2-shot scenario, during inference it can operate for any number of source frames $N$. We examine the setting where $N=2,4$ or 8 source frames are given and report the results on the task of reconstruction of VoxCeleb test set, in Table~\ref{table:N-shot}. As suggested by all evaluation metrics, the generative performance of our system improves when the number of available images increases. In Fig.~\ref{fig:shot} we present visual examples that show the effect of the number of source images $N$ in the quality of synthesised data. The results clearly demonstrate the beneficial effect of N-shot learning.

Next, we examine the significance of the two front-end components of our talking head synthesis system and more specifically networks $E_{can}$ and $E_{gaze}$.

We focus on cross-identity motion transfer, as the canonical key-point estimator $E_{can}$ is particularly developed to tackle the identity mismatches in reenactment, since it allows to adapt the target key-points to the facial shape of the source. In order to evaluate its contribution, we develop a variation of \textit{Free-HeadGAN} by replacing $E_{can}$ network with \textit{RetinaFace} \cite{RetinaFace}. That is, we utilise directly the 3D key-points regressed by \textit{RetinaFace} in order to draw the sketches that serve as conditional input to the generator, without adapting them to the identity of the source. The quantitative results displayed in Table~\ref{table:ablation} indicate that removing $E_{can}$ has a critical effect on CSIM metric that measures identity preservation. Moreover, the overall quality of samples degrades as FID score increases. On the contrary, expression transfer slightly improves, indicating that the canonical space key-points actually retain a small amount of expression information.

\begin{figure}[!t]
\centering
\begin{picture}(260,185)
\put(0,20){
\includegraphics[width=0.48\textwidth]{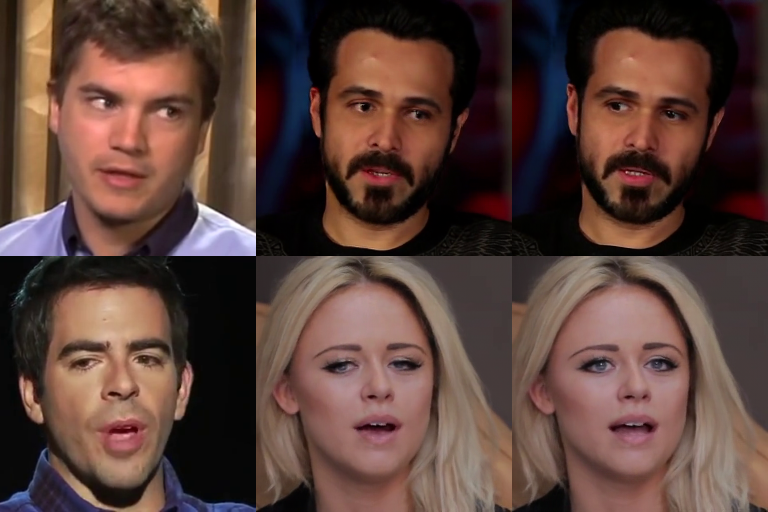}}
\put(30,10){target}
\put(100,10){\textit{Free-HeadGAN}}
\put(185,10){\textit{Free-HeadGAN}}
\put(195,0){w/o $E_{gaze}$}
\end{picture}
\caption{Examples where explicit gaze control is essential for successful gaze transfer, as 68 facial landmarks appear to be insufficient.}
\label{fig:gaze}
\end{figure}
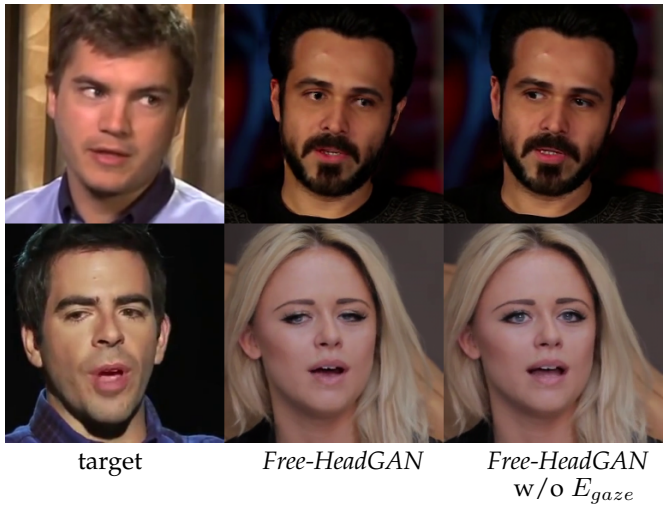

In order to validate the importance of explicit eye gaze conditioning in image synthesis, we test a second variation of our system, where we do not encode any gaze information into key-point sketches. To that end, we train a \textit{Free-HeadGAN} generator that learns to transfer gaze direction relying solely on the facial key-points that belong to the eyes, as we do not color code the gaze angles inside the eye cavities. We evaluate the performance of this variation with AGD metric, which measures the average angular distance between the driving and generated gaze on the test set for reenactment. We display AGD results in Table~\ref{table:ablation}, which confirms the significance of gaze estimator $E_{gaze}$ quantitatively. Additionally, in Fig.~\ref{fig:gaze} we illustrate some cases where the model variation without explicit gaze input fails to capture the target eyes direction.

Lastly, we quantitatively verify the benefit of implicitly inferring gaze from predicted dense 3D eye coordinates instead of learning to directly predict few gaze parameters. To that end, we train two versions of $E_{gaze}$ one predicting dense eye meshes as described in Sec.~\ref{subsec:gaze}, $E_{gaze}^{meshes}$, and one predicting 3D gaze vectors, $E_{gaze}^{vectors}$. We perform within-dataset, cross-subject experiments on the recent gaze estimation datasets MPIIGaze~\cite{zhang15_cvpr_mpiigaze}, Columbia~\cite{CAVE_Columbia2013}, UTMV~\cite{sugano2014utmv}, and Gaze360~\cite{gaze360_2019}. We compute the AGD and present results in Table~\ref{table:gaze}, which confirms our initial suggestion. 

\begin{table}[!h]
    \setlength{\tabcolsep}{6pt}
    \renewcommand{\arraystretch}{1.15}
    \centering
    \caption{Comparison between the 3D mesh and 3D vector regression approaches to gaze estimation, on within-dataset, cross-subject experiments.}
    \label{table:gaze}
    \begin{tabular}{|c||c|c|c|c|}
    \hline
     & \multicolumn{4}{c|}{\textbf{AGD} $\downarrow$} \\
    \hline
    Variation & MPIIGaze & Columbia & UTMV & Gaze360 \\
    \hline
    $E_{gaze}^{vectors}$ & 4.83 & 3.84 & 5.62 & 12.7 \\
    $E_{gaze}^{meshes}$ & 3.35 & 1.12 & 4.11 & 10.4 \\
    \hline
    \end{tabular}
\end{table}

\subsection{Pose and Gaze Manipulation}

\begin{figure}[!t]
\centering
\includegraphics[width=0.485\textwidth]{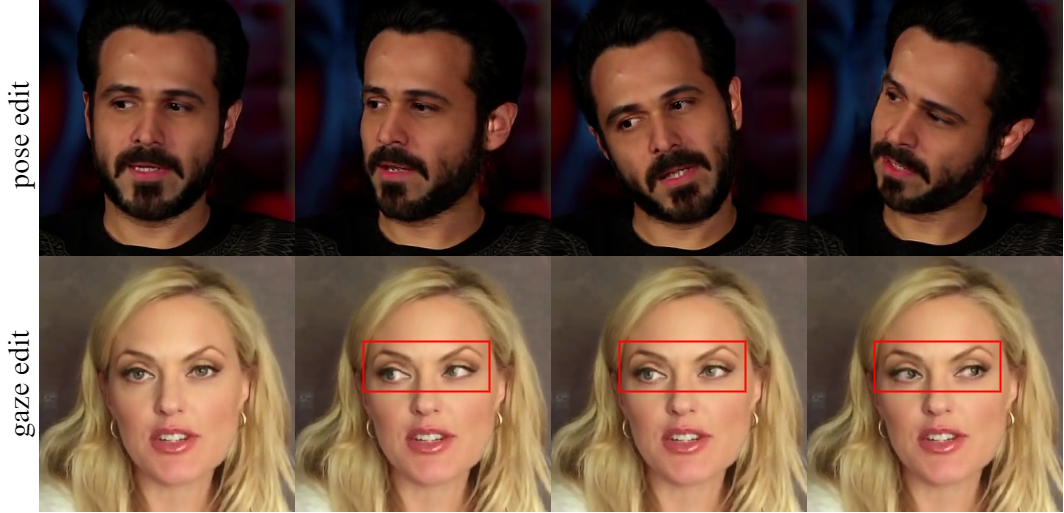}
\caption{The 3D key-point modeling of faces enables our system to perform free-view synthesis, as we are able to manually edit the head pose of a given reference image. Furthermore, we have explicit gaze control over the subject.}
\label{fig:edit}
\end{figure}

Apart from cross-identity motion transfer (reenactment) and facial video compression and reconstruction (self-reenactment) we can use \textit{Free-HeadGAN} to edit portrait images. That is, we can set the driving head pose manually, simply by rotating the 3D source key-points. Moreover, we can change the gazing direction of the reference subject, by editing the estimated gaze angles before feeding them to the generator through the target sketch. In Fig.~\ref{fig:edit} we demonstrate the ability of our proposed system to edit the head pose and eye gaze of any given reference image. Here, the left column shows the original reference image, while the next three columns depict synthesised images  by our method, after editing pose and gaze. Our image editing task makes \textit{Free-HeadGAN} a powerful tool for data augmentation, as our method can replace the naive and widely-used affine transformations on image data with complex non-linear transformations of human heads. Moreover, considering that our system provides strong identity preservation, such image augmentation can benefit various computer vision tasks related to face recognition.

\section{Limitations}

Although our method performs one-shot reenactment with unparalleled image quality and photo-realism while proving explicit gaze control and strong identity preservation properties, it does not come without any limitations. Considering that high quality data are crucial for deep learning algorithms, the same applies to our system. In particular, \textit{Free-HeadGAN}'s performance in extreme target poses, such as large head rotations, highly depends on the quite limited pose distribution in open source databases such as \cite{voxceleb}. Moreover, as suggested by our qualitative analysis and FID scores \cite{NIPS2017_8a1d6947}, there exists a significant performance gap between self-reenactment and reenactment. Aside from the adaptation of key-points in cross-identity motion transfer, this gap is caused by the random selection of source and target image pairs during training and could be improved by a more sophisticated learning strategy, where progressively tougher pairs are chosen. %Another shortcoming of our approach is associated with expression modeling and disentanglement from identity-related components in facial landmarks. Please refer to Sec. \ref{sec:comparison} for a more detailed discussion on this topic. 
Last but not least, even though our method does not rely on 3DMMs \cite{3dmms, 3dmmsbooth, Booth, Booth_2016_CVPR}, the supervision of our networks depends on the quality of 3D landmark pseudo annotations with \textit{RetinaFace} \cite{RetinaFace} and labeled gaze data, which come with their own inaccuracies.

\section{Conclusion}

We presented \textit{Free-HeadGAN}, a model that extends \textit{HeadGAN}~\cite{doukas2020headgan} while releasing it from 3DMM fitting and 3D face rendering requirements in pre-processing. Instead, we condition synthesis on 3D landmarks which also support free-view synthesis. We designed a network that takes care of the identity mismatches in cross-identity motion transfer. Finally, we proposed a gaze estimation module and demonstrated the advantage on utilising gaze signals for precisely controlling of eye movements of the generated samples. 

\appendix  % for no appendix heading

\section*{Architecture Details}

\begin{figure}[h!]
\includegraphics[scale=0.69]{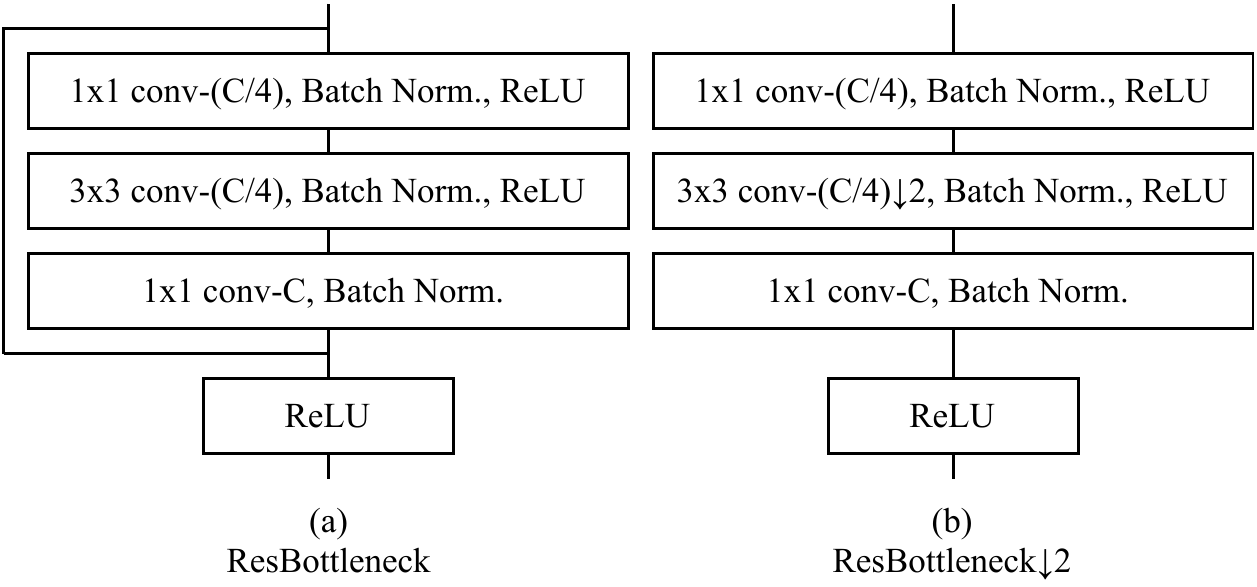}
\caption{(a) The ResBottleneck. (b) The ResBottleneck that downsamples the input tensor (stride$=2$) does not have a residual component. Here, C denotes the number of output channels.}
\label{fig:fig1}
\end{figure}

\textbf{Network $E_{can}$.} This network receives as input an image and regresses: a) an affine transformation (scale, rotation, translation), b) the expression-related deformation and c) the 68 3D key-points. The architecture of this module is inspired from that of the head pose estimator and expression deformation estimator in \cite{wang2021facevid2vid}. We present it in Table~\ref{table:1}. Please note that for rotation we do not predict Euler degrees directly but compute the Softmax over $2 n_{deg} + 1$ values, corresponding to degrees in the interval $\{- n_{deg} .. n_{deg}\}$, with $n_{deg}=60$. The ResBottleneck block architecture is depicted in Fig~\ref{fig:fig1}.

\begin{table}[h!]
\centering
\renewcommand{\arraystretch}{1.15}
\begin{tabular}{c c c | c} 
\hline
& Block &  & Output size \\
\hline\hline
& Input image&  & $(256, 256, 3)$ \\
$7 \times 7$ conv-32 $\downarrow2$ & Batch Norm. & ReLU & $(128, 128, 32)$ \\ 
\hline
& ResBottleneck $\downarrow2$ &  & $(64, 64, 128)$ \\
& ResBottleneck $(\times 3)$ &  & $(64, 64, 128)$ \\
& ResBottleneck $\downarrow2$ &  & $(32, 32, 256)$ \\
& ResBottleneck $(\times 3)$ &  & $(32, 32, 256)$ \\
& ResBottleneck $\downarrow2$ &  & $(16, 16, 512)$ \\
& ResBottleneck $(\times 5)$ &  & $(16, 16, 512)$ \\
& ResBottleneck $\downarrow2$ &  & $(8, 8, 1024)$ \\
& ResBottleneck $(\times 2)$ &  & $(8, 8, 1024)$ \\
& $8 \times 8$ AvgPool2d & & $1024$ \\
\hline
& Affine $1024 \times 121$  & Softmax & $1$ (pitch) \\
& Affine $1024 \times 121$  & Softmax & $1$ (yaw) \\
& Affine $1024 \times 121$  & Softmax & $1$ (roll) \\
& Affine $1024 \times 3$  & & $3$ (translation) \\
& Affine $1024 \times 1$  & & $1$ (scale) \\
& Affine $1024 \times 68 \cdot 3$  & & $68 \cdot 3$ (expr.) \\
& Affine $1024 \times 68 \cdot 3$  & & $68 \cdot 3$ (points) \\
\hline
\end{tabular}
\caption{Architecture of network $E_{can}$.}
\label{table:1}
\end{table}

\textbf{Network $E_{gaze}$.}v The gaze estimation network $E_{gaze}$ takes as input a square image cropped around the eye, resized to resolution $128 \times 128$ and predicts $N_{v}^{eye}$ vertices that correspond to the mesh of the eye. Table~\ref{table:2} shows the architecture of $E_{gaze}$, which is based on ResNet-34.

\begin{table}[h!]
\centering
\renewcommand{\arraystretch}{1.15}
\begin{tabular}{c c c | c} 
\hline
& Block &  & Output size \\
\hline\hline
& Input eye image &  & $(128, 128, 3)$ \\
\hline
& ResNet-34 &  & $(4, 4, 512)$ \\
& $3 \times 3$ conv-512 $\downarrow 2$ &  & $(2, 2, 512)$ \\
\hline
& Affine $2048 \times N_{v}^{eye} \cdot 3$  & & $N_{v}^{eye} \cdot 3$ (eye mesh) \\
\hline
\end{tabular}
\caption{Architecture of network $E_{gaze}$. Here $N_{v}^{eye} = 481$.}
\label{table:2}
\end{table}

\noindent \textbf{Flow network} $F$ (Table \ref{table:3}). The flow network consists of an encoder and a decoder, as presented originally in HeadGAN~\cite{doukas2020headgan}. The encoder has three convolutional layers, each one with instance normalization (IN) units \cite{instancenorm} and ReLU activation layers. The last convolutions are performed with a stride of 2, for down-sampling the input. The SPADE blocks \cite{park2019SPADE} of the decoder are used as modulation inputs for injecting the driving (target) sketch $\textbf{x}^{t}$. Similarly with SPADE \cite{park2019SPADE}, we down-sample $\textbf{x}^{t}$ to match the spatial size of each SPADE block. As in \cite{doukas2020headgan}, we employ two Pixel Shuffle \cite{pixelshuffle} layers for up-sampling. The dense optical flow is computed with a $7 \times 7$ convolutional layer. For the extension of our model to perform N-shot learning, we define an extra output layer, that estimates the weights. 

\begin{table}[t!]
\centering
\renewcommand{\arraystretch}{1.15}
\begin{tabular}{c c c | c} 
\hline
& Block &  & Output size \\
\hline\hline
& Input &  & $(256, 256, 6)$ \\
$7 \times 7$ conv-32 & Inst. Norm. & ReLU & $(256, 256, 32)$ \\ 
$3 \times 3$ conv-128 & Inst. Norm. & ReLU & $(128, 128, 128)$ \\
$3 \times 3$ conv-512 & Inst. Norm. & ReLU & $(64, 64, 512)$ \\
\hline
& SPADE Block &  & $(64, 64, 512)$ \\
& SPADE Block &  & $(64, 64, 512)$ \\
& SPADE Block &  & $(64, 64, 512)$ \\
& Pixel Shuffle &  & $(128, 128, 128)$ \\
& SPADE Block &  & $(128, 128, 128)$ \\
& Pixel Shuffle &  & $(256, 256, 32)$ \\
\hline
& $7 \times 7$ conv-2  &  & $(256, 256, 2)$ \\
& $7 \times 7$ conv-1 & $\tanh$ & $(256, 256, 1)$ \\
\hline
\end{tabular}
\caption{Architecture of flow network $F$.}
\label{table:3}
\end{table}

\noindent \textbf{Rendering network} $R$ (Table \ref{table:4}). The rendering network has an architecture similar with the one of HeadGAN~\cite{doukas2020headgan}. The encoder has exactly the same layers with the encoder of the rendering network in \cite{doukas2020headgan}. On the contrary, the decoder is equipped only with SPADE blocks, which are used to condition synthesis on the warped visual feature maps and the warped source image. Here, we do not include AdaIN layers \cite{adain}, as we do not condition image synthesis on audio features. Pixel Shuffle layers \cite{pixelshuffle} are used for up-sampling. After the final decoding block, a convolutional layer hallucinates of the generated image.

\begin{table}[h!]
\centering
\renewcommand{\arraystretch}{1.15}
\begin{tabular}{c c c | c} 
\hline
& Block &  & Output size \\
\hline\hline
& Input &  & $(256, 256, 9)$ \\
$7 \times 7$ conv-32 & Inst. Norm. & ReLU & $(256, 256, 32)$ \\ 
$3 \times 3$ conv-128 & Inst. Norm. & ReLU & $(128, 128, 128)$ \\
$3 \times 3$ conv-512 & Inst. Norm. & ReLU & $(64, 64, 512)$ \\
\hline
& SPADE Block &  & $(64, 64, 512)$ \\
& Pixel Shuffle &  & $(128, 128, 128)$ \\
& SPADE Block &  & $(128, 128, 128)$ \\
& Pixel Shuffle &  & $(256, 256, 32)$ \\
& SPADE Block &  & $(256, 256, 32)$ \\
& SPADE Block &  & $(256, 256, 32)$ \\
\hline
LReLU & $7 \times 7$ conv-3 & $\tanh$ & $(256, 256, 3)$ \\
\hline
\end{tabular}
\caption{Architecture of rendering network $R$.}
\label{table:4}
\end{table}

\textbf{Discriminators $D$ and $D_m$}. The image and mouth discriminators $D$ and $D_m$ have a similar architecture to the discriminator presented in \cite{doukas2020headgan} and \cite{park2019SPADE}. 

\section*{Additional results}

In Fig.~\ref{fig:fig3} to \ref{fig:fig8} we illustrate more generated samples with Free-HeadGAN model on VoxCeleb test set \cite{voxceleb}.

\begin{figure*}[!ht]
\centering
\begin{picture}(520,330)
\put(0,10){\includegraphics[width=0.98\textwidth]{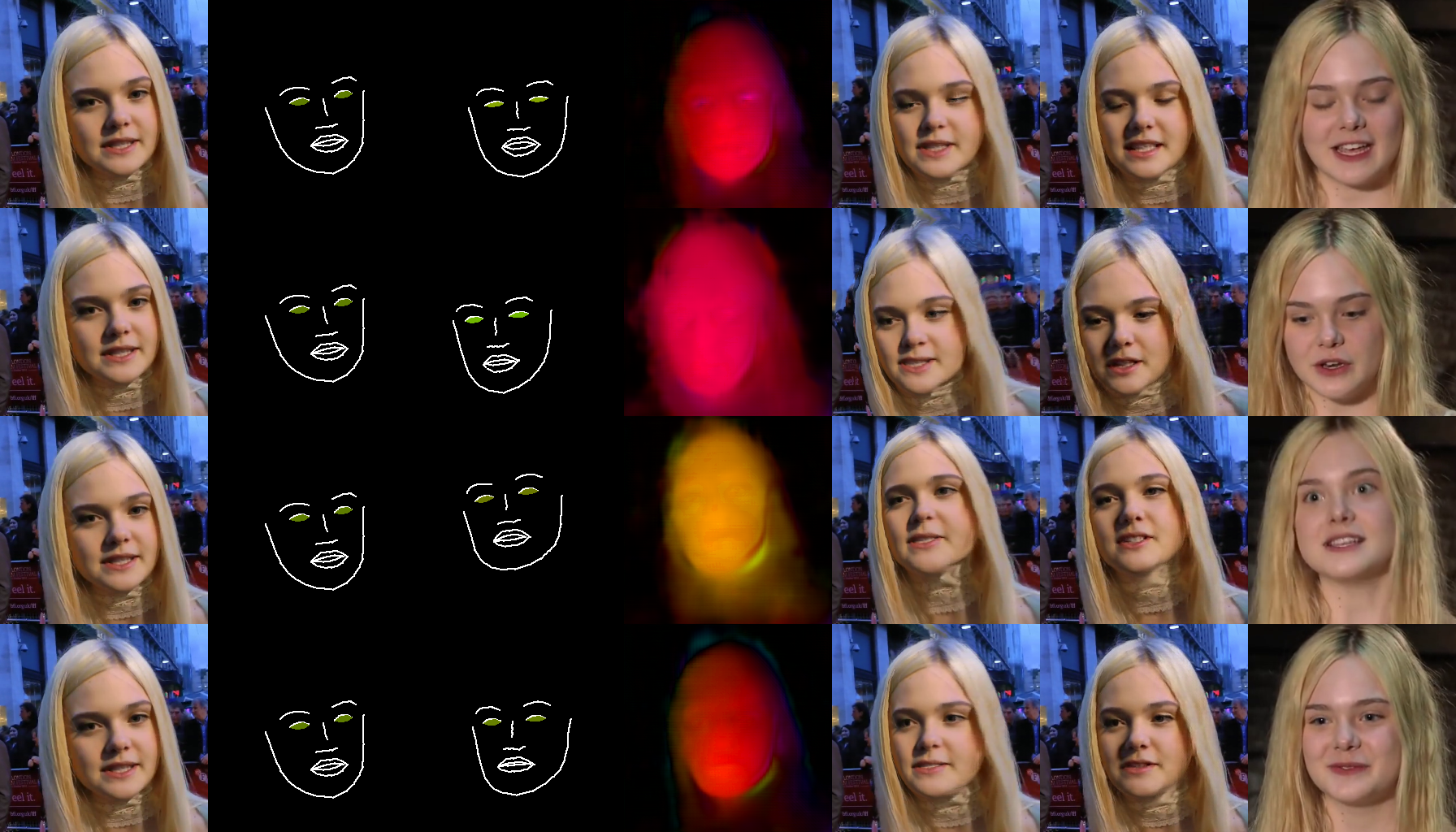}}
\put(24,0){source}
\put(80,0){source sketch}
\put(155,0){target sketch}
\put(230,0){optical flow}
\put(310,0){warped}
\put(375,0){generated}
\put(455,0){target}
\end{picture}
\caption{Reenactment results (a)}
\label{fig:fig3}

\end{figure*}
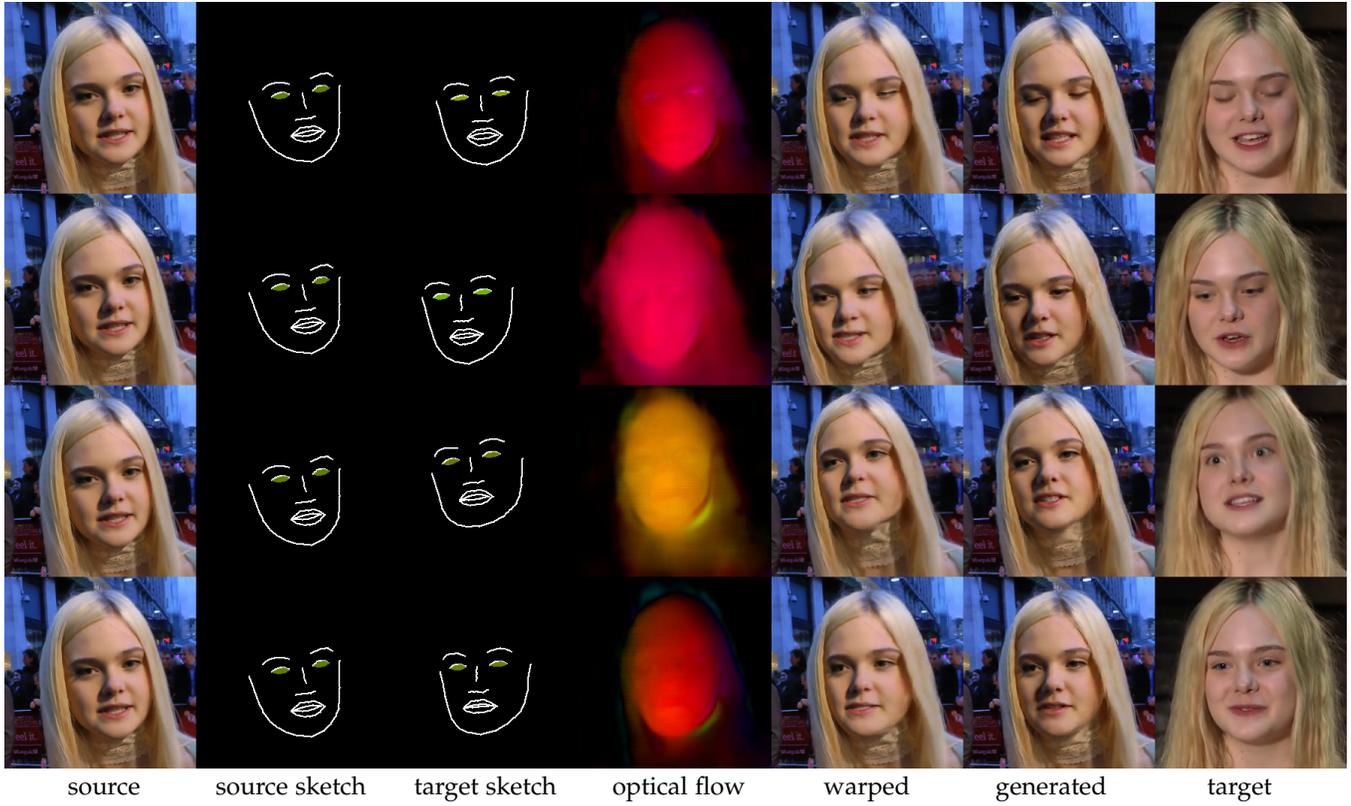
\begin{figure*}[!hb]
\centering
\begin{picture}(520,300)
\put(0,10){\includegraphics[width=0.98\textwidth]{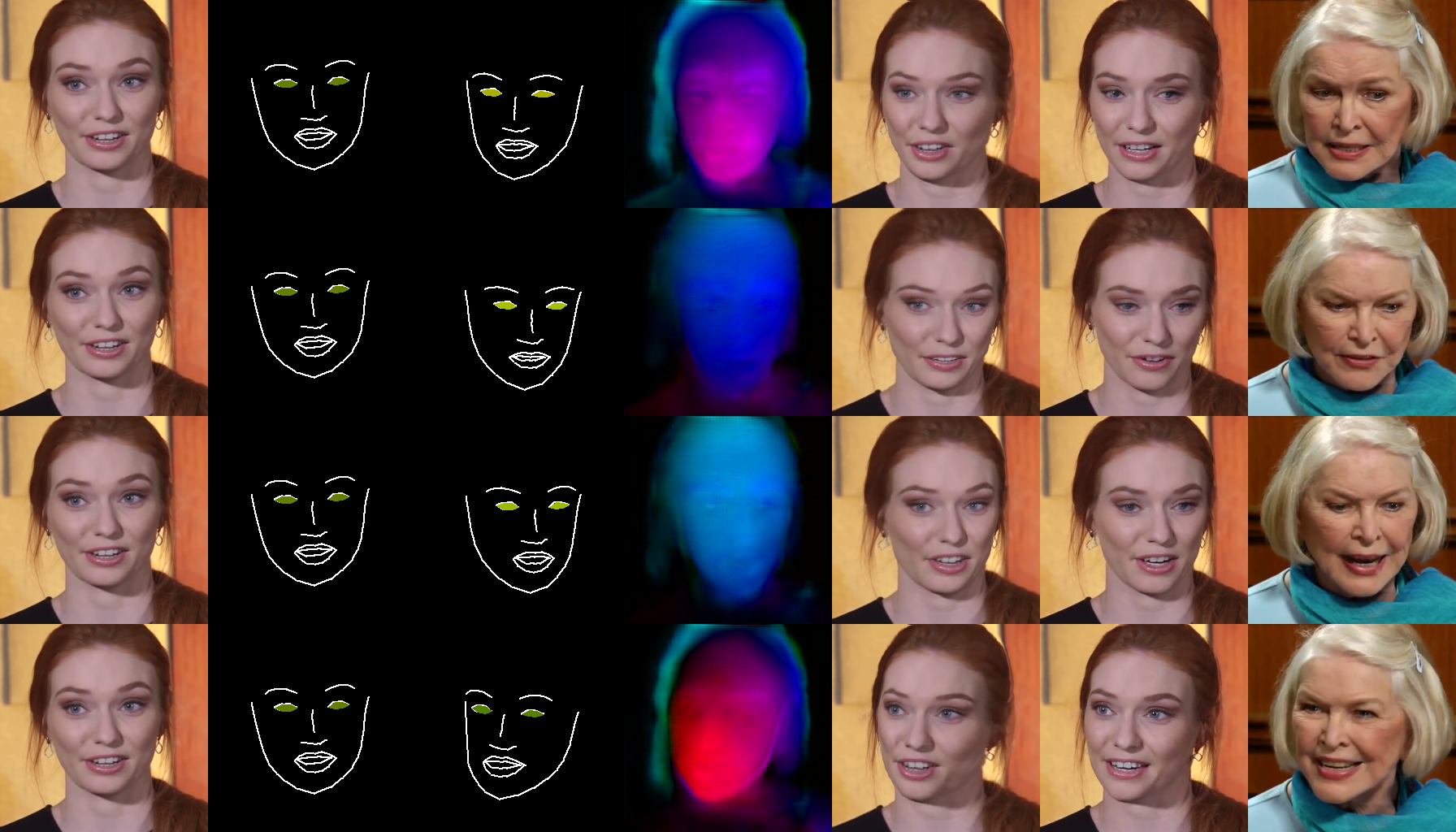}}
\put(24,0){source}
\put(80,0){source sketch}
\put(155,0){target sketch}
\put(230,0){optical flow}
\put(310,0){warped}
\put(375,0){generated}
\put(455,0){target}
\end{picture}
\caption{Reenactment results (b)}
\label{fig:fig4}
\end{figure*}

\begin{figure*}[!ht]
\centering
\begin{picture}(520,330)
\put(0,10){\includegraphics[width=0.98\textwidth]{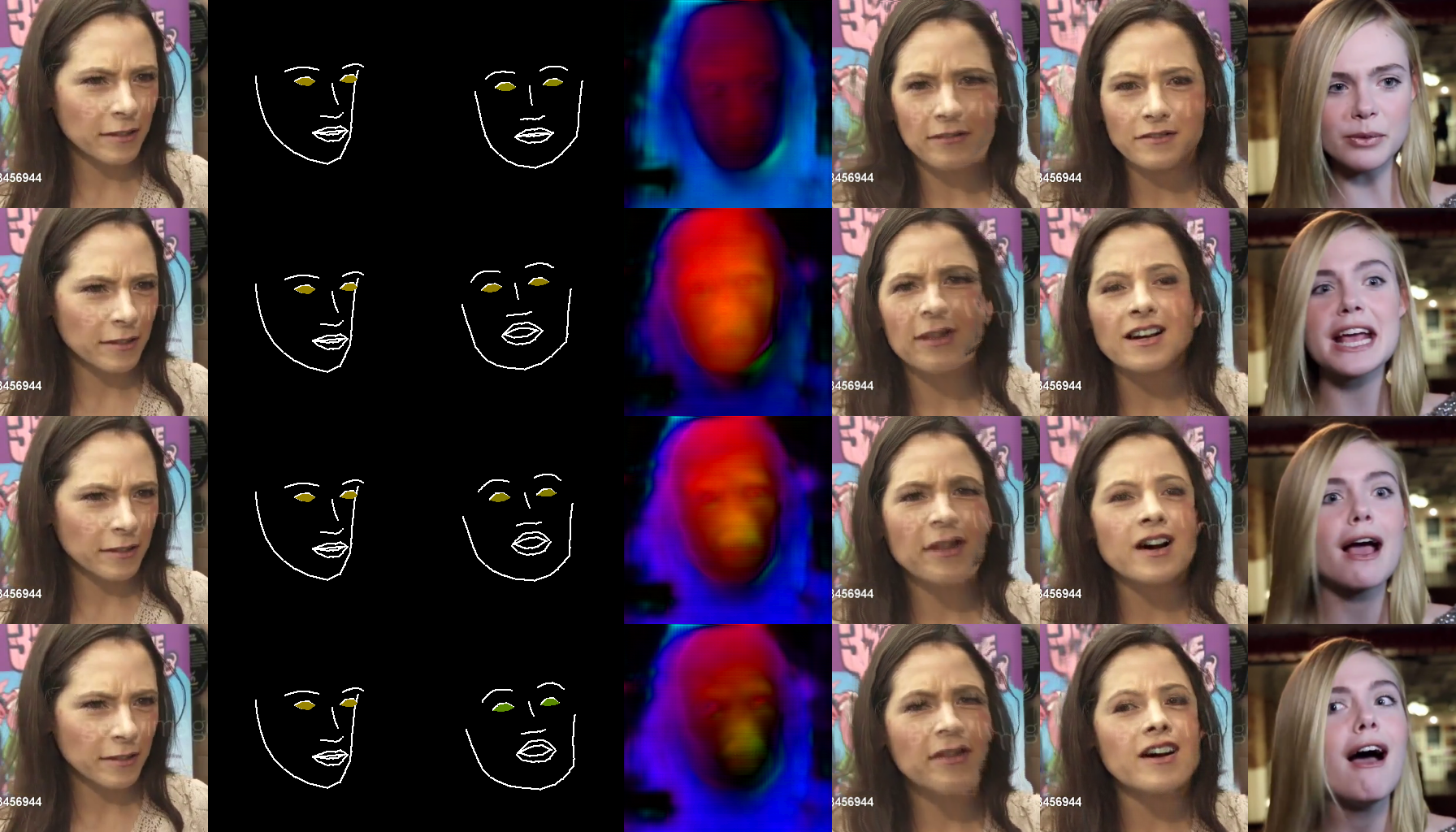}}
\put(24,0){source}
\put(80,0){source sketch}
\put(155,0){target sketch}
\put(230,0){optical flow}
\put(310,0){warped}
\put(375,0){generated}
\put(455,0){target}
\end{picture}
\caption{Reenactment results (c)}
\label{fig:fig5}
\end{figure*}

\begin{figure*}[!hb]
\centering
\begin{picture}(520,300)
\put(0,10){\includegraphics[width=0.98\textwidth]{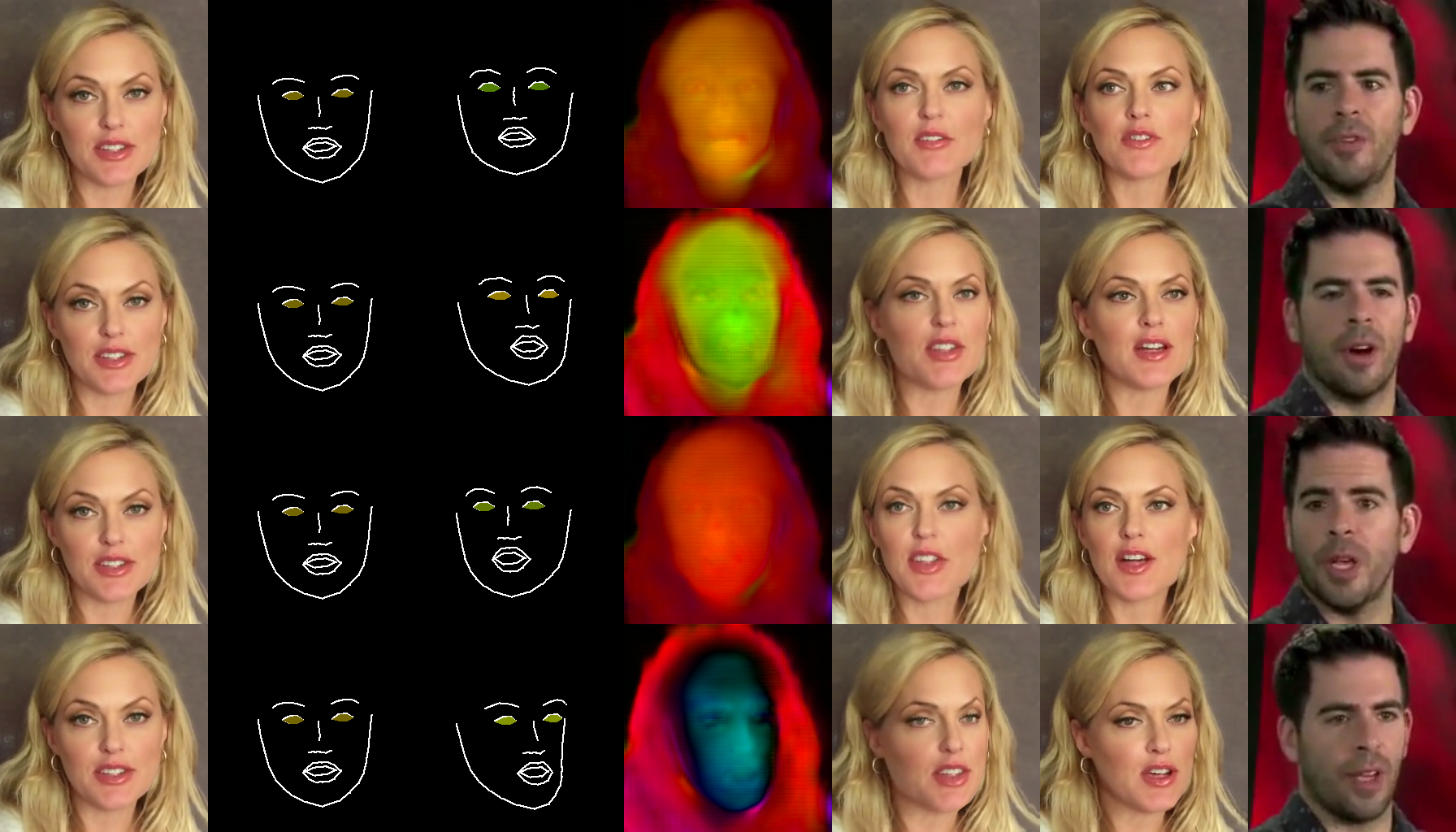}}
\put(24,0){source}
\put(80,0){source sketch}
\put(155,0){target sketch}
\put(230,0){optical flow}
\put(310,0){warped}
\put(375,0){generated}
\put(455,0){target}
\end{picture}
\caption{Reenactment results (d)}
\label{fig:fig6}
\end{figure*}

\begin{figure*}[!ht]
\centering
\begin{picture}(520,330)
\put(0,10){\includegraphics[width=0.98\textwidth]{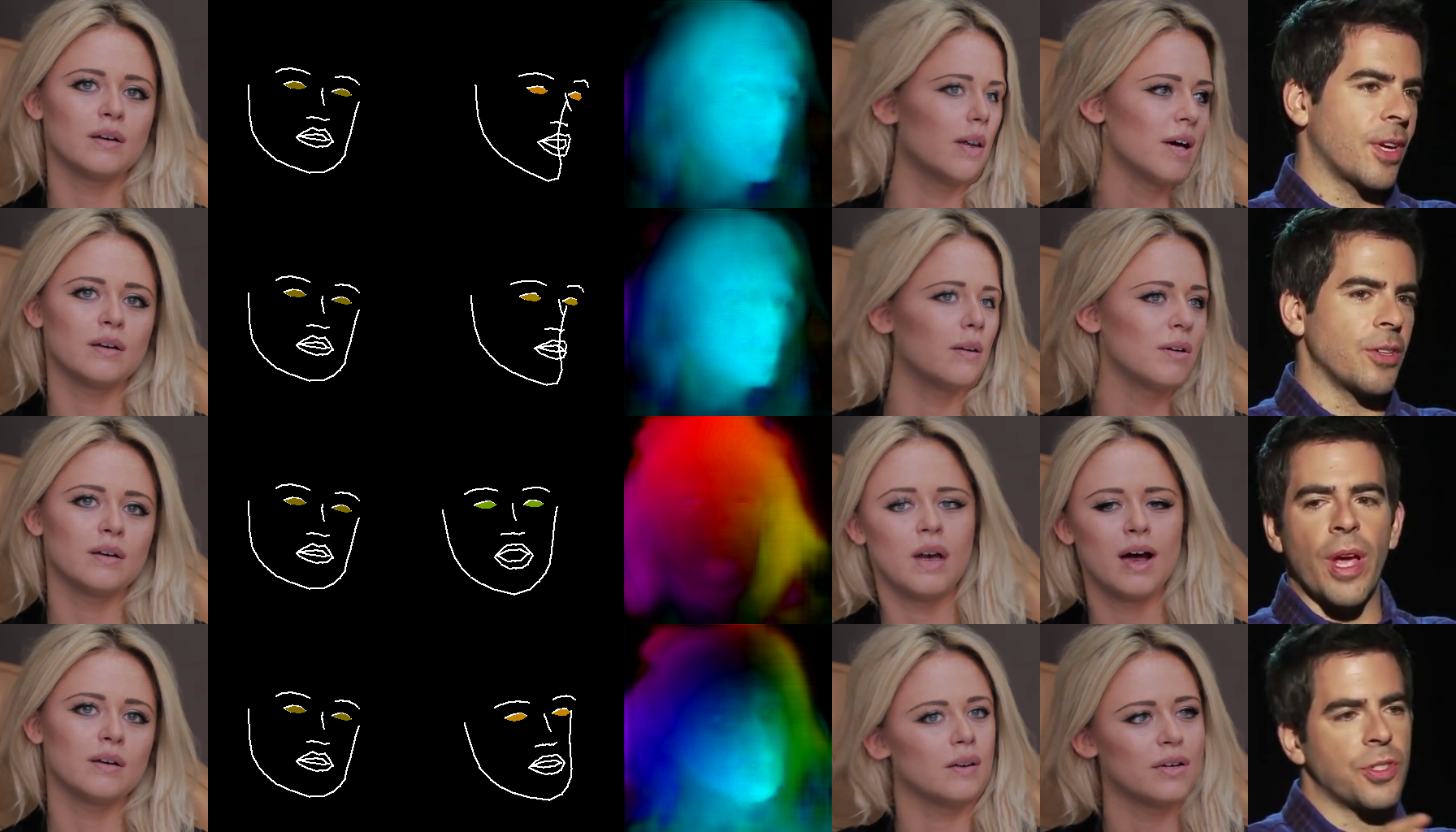}}
\put(24,0){source}
\put(80,0){source sketch}
\put(155,0){target sketch}
\put(230,0){optical flow}
\put(310,0){warped}
\put(375,0){generated}
\put(455,0){target}
\end{picture}
\caption{Reenactment results (e)}
\label{fig:fig7}

\end{figure*}
\begin{figure*}[!hb]
\centering
\begin{picture}(520,300)
\put(0,10){\includegraphics[width=0.98\textwidth]{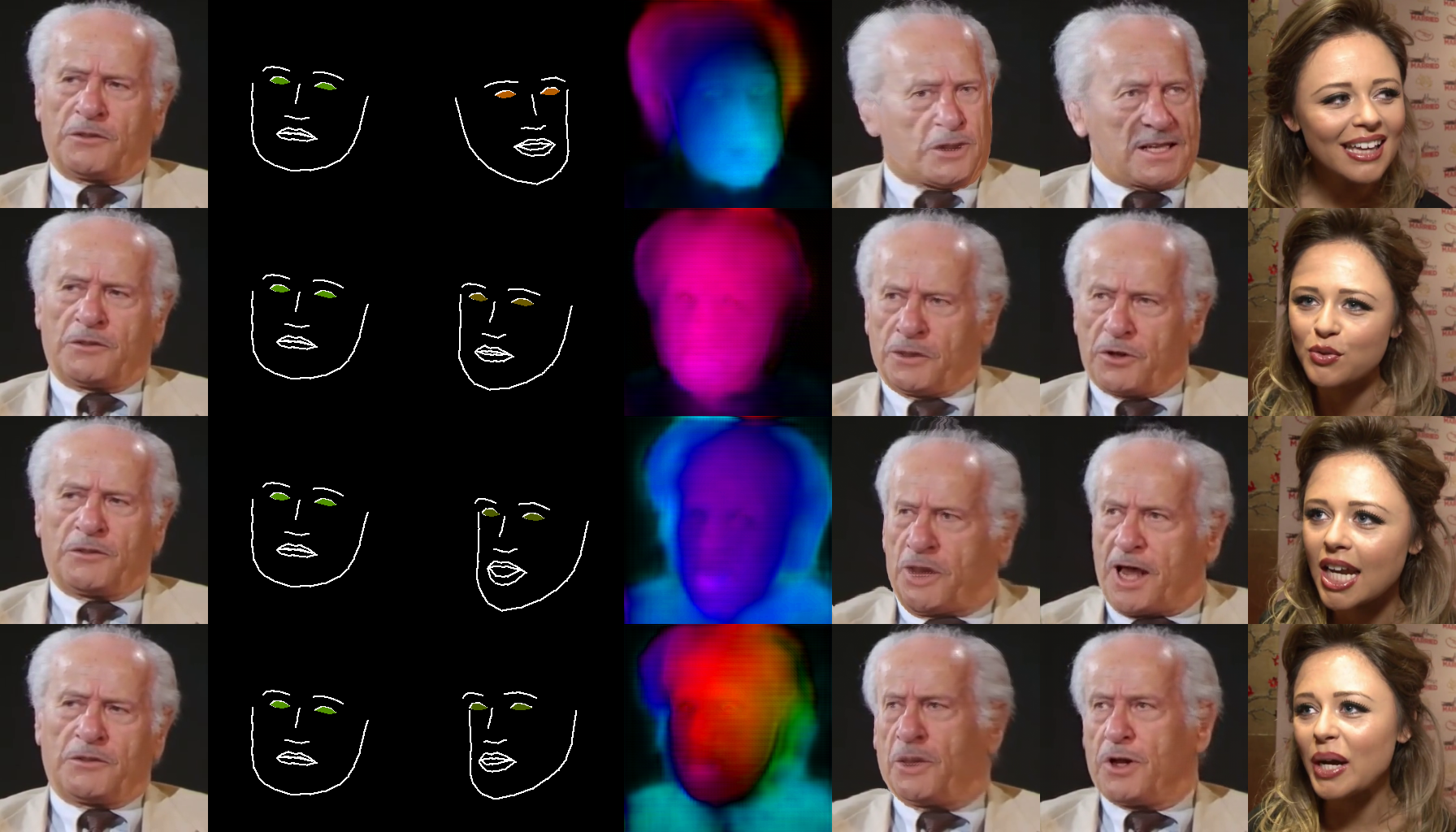}}
\put(24,0){source}
\put(80,0){source sketch}
\put(155,0){target sketch}
\put(230,0){optical flow}
\put(310,0){warped}
\put(375,0){generated}
\put(455,0){target}
\end{picture}
\caption{Reenactment results (f)}
\label{fig:fig8}
\end{figure*}


%\ifCLASSOPTIONcompsoc
%\else
%\fi

\ifCLASSOPTIONcaptionsoff
  \newpage
\fi


\bibliographystyle{IEEEtran}
\bibliography{Free-HeadGAN.bib}


\begin{IEEEbiography}[{\includegraphics[width=1in,height=1.25in,clip,keepaspectratio]{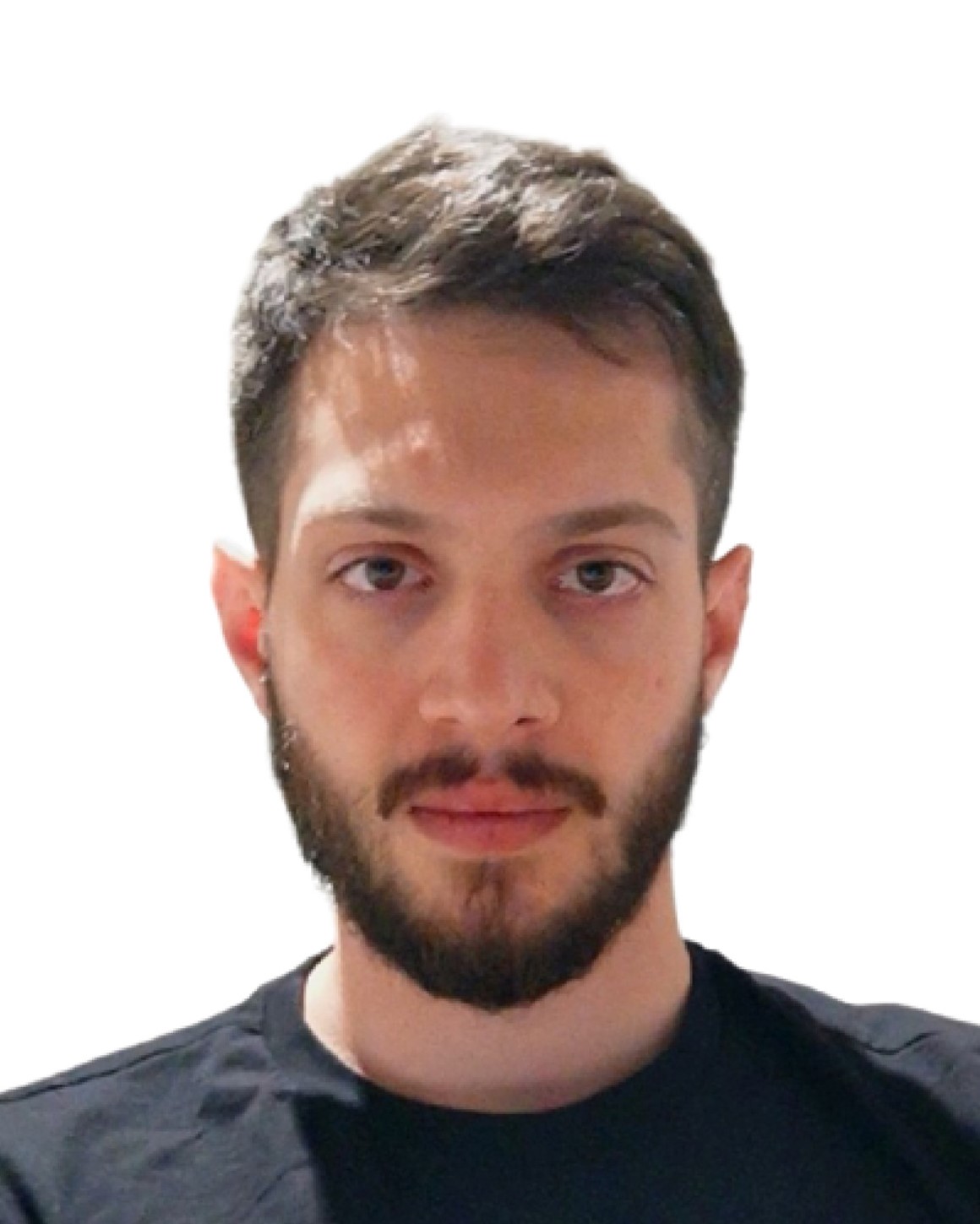}}]{Michail Christos Doukas}
is a 4th year PhD Student at Imperial College London, supervised by Dr. Viktoriia Sharmanska and Prof. Stefanos Zafeiriou. He is currently an intern at Huawei UK Research Centre. He received the MSc degree in Computing from Imperial College London, in 2017. Prior to that, he has studied Electrical and Computer Engineering (MEng 2016) at the National Technical University of Athens (NTUA), Greece. His research interests lie in the fields of Deep Learning and Computer Vision and include generative adversarial neural networks, image and video synthesis, visual speech synthesis, face reenactment and few-shot learning.
\end{IEEEbiography}

\begin{IEEEbiography}[{\includegraphics[width=1in,height=1.25in,clip,keepaspectratio]{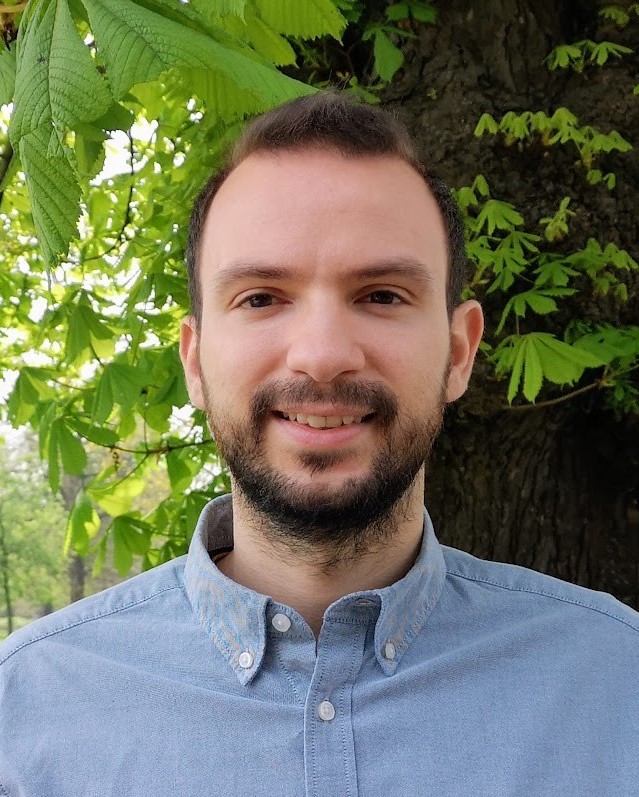}}]{Evangelos Ververas} graduated in September 2016 from the Department of Electrical and Computer Engineering in Aristotle University of Thessaloniki (AUTH), in Greece. He is currently a PhD student at the Department of Computing at Imperial College London, supervised by Prof. Stefanos Zafeiriou, and a researcher at Huawei UK Research Centre. His research interests lie in the fileds of Computer Vision and Deep Learning for 3D face analysis and reconstruction, as well as 3D gaze estimation.
\end{IEEEbiography}

\begin{IEEEbiography}[{\includegraphics[width=1in,height=1.25in,clip,keepaspectratio]{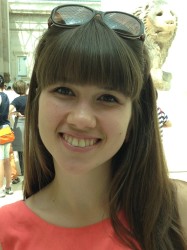}}]{Viktoriia Sharmanska}
Viktoriia Sharmanska is passionate about designing intelligent systems that can learn concepts from visual data using machine learning models. She has joined the University of Sussex as a Lecturer in Artificial Intelligence in 2020. Prior to this, 2017-2020, she was an Imperial College Research Fellow at the Department of Computing, Imperial College London, working on deep learning methods for emotion and human behaviour recognition. She is also an honorary lecturer at Imperial College London since 2021. Dr Sharmanska has co-authored numerous papers on novel statistical machine learning methodologies applied to computer vision problems, such as attribute-based object recognition, learning using privileged information, cross-modal learning, human facial behaviour analysis, and recently on algorithmic fairness methods, published in the most prestigious conferences in her field of research, such as CVPR, ICCV/ECCV, NeurIPS. She has built an international reputation such as being among the youngest Area Chair for top-tier international conferences in computer vision and deep learning such as International Conference on Learning Representations (ICLR) 2019-2021 and Conference on Computer Vision and Pattern Recognition (CVPR) 2021. Viktoriia is actively involved in organising and promoting Women in Computer Vision, Women in Machine Learning, and in increasing visibility of the underrepresented groups in computing. In 2018, together with colleagues from Facebook AI, Stanford University, Autonomous University of Barcelona she co-organised the international Women in Computer Vision Workshop, held in conjunction with CVPR2018.
\end{IEEEbiography}

%\newpage

\begin{IEEEbiography}[{\includegraphics[width=1in,height=1.25in,clip,keepaspectratio]{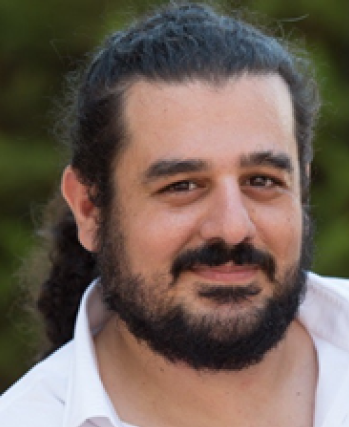}}]{Stefanos Zafeiriou}
(M’09) is a Reader in Machine
Learning and Computer Vision with the Department
of Computing, Imperial College London,
U.K, and a Distinguishing Research Fellow with
University of Oulu. He was a recipient of the Prestigious
Junior Research Fellowships from Imperial
College London in 2011 to start his own independent
research group. He was the recipient of
the President’s Medal for Excellence in Research
Supervision for 2016. He currently serves as an
Associate Editor of the IEEE Transactions on
Affective Computing and Computer Vision and Image Understanding journal.
He has been a Guest Editor of over six journal special issues and coorganised
over 13 workshops/special sessions on specialised computer
vision topics in top venues, such as CVPR/FG/ICCV/ECCV. He has coauthored
over 55 journal papers mainly on novel statistical machine learning
methodologies applied to computer vision problems, such as 2-D/3-D
face analysis, deformable object fitting and tracking, published in the most
prestigious journals in his field of research, such as the IEEE T-PAMI,
the International Journal of Computer Vision, the IEEE T-IP, the IEEE
T-NNLS, the IEEE T-VCG, and the IEEE T-IFS, and many papers in top
conferences. He has more than 18000 citations to his work, h-index 51.
\end{IEEEbiography}


%\vfill


\end{document}